\documentclass{article}
\usepackage{amssymb}

\PassOptionsToPackage{table}{xcolor}\usepackage{tcolorbox}
\PassOptionsToPackage{numbers,sort&compress}{natbib}

\usepackage[final]{corl_2025} %

\usepackage{multicol}
\usepackage{xspace}
\usepackage{graphicx}
\usepackage{float}
\usepackage{xcolor}
\usepackage{siunitx}
\usepackage{mathtools}
\usepackage{amsfonts}
\usepackage{booktabs}
\usepackage{graphicx}
\usepackage{amsmath}
\usepackage{caption}
\usepackage{subcaption}
\usepackage{amssymb}
\usepackage{adjustbox}
\usepackage{listings}
\usepackage{multirow}
\usepackage{wrapfig}

\usepackage{fontawesome}
\newcolumntype{R}[2]{%
    >{\adjustbox{angle=#1,lap=\width-(#2)}\bgroup}%
    l%
    <{\egroup}%
}
\newcommand*\rot{\multicolumn{1}{R{45}{1em}}}

\usepackage{rotating}
\definecolor{forestgreen}{rgb}{0.13, 0.55, 0.13}
\usepackage{threeparttable}

\definecolor{codegreen}{rgb}{0,0.6,0}
\definecolor{codegray}{rgb}{0.5,0.5,0.5}
\definecolor{codepink}{RGB}{252, 142, 172}
\definecolor{codepurple}{rgb}{0.58,0,0.82}
\definecolor{backcolour}{RGB}{245,245,245}
\lstdefinestyle{mystyle}{
    backgroundcolor=\color{backcolour},   
    commentstyle=\color{magenta},
    keywordstyle=\color{blue},
    numberstyle=\tiny\color{codegray},
    stringstyle=\color{codepurple},
    basicstyle=\fontfamily{\ttdefault}\footnotesize,
    breakatwhitespace=false,         
    breaklines=true,                 
    captionpos=b,                    
    keepspaces=true,    
    frame=single,
    numbersep=5pt,                  
    showspaces=false,                
    showstringspaces=false,
    showtabs=false,                  
    tabsize=2,
    classoffset=1, %
    otherkeywords={range},
    keywordstyle=\color{violet},
    classoffset=0,
}

\definecolor{mobility_color}{HTML}{F0C987}
\newcommand\mobility[1]{\textcolor{mobility_color}{{\fontseries{b}\selectfont #1}}}
\definecolor{bimanual_color}{HTML}{92C4E9}
\newcommand\bimanual[1]{\textcolor{bimanual_color}{{\fontseries{b}\selectfont #1}}}
\definecolor{reachability_color}{HTML}{EA9A9D}
\newcommand\reachability[1]{\textcolor{reachability_color}{{\fontseries{b}\selectfont #1}}}

\definecolor{bestscorecolor}{RGB}{249,137,72}
\definecolor{cardinal}{RGB}{140,21,21}

\newcommand{\bestscore}[1]{\textcolor{bestscorecolor}{\textbf{#1}}}

\newcommand{\para}[1]{\paragraph{#1}\looseness=-1}
\newcommand{\q}[1]{$\mathbf{\mathcal{Q}#1}$}
\newcommand{\webpage}[0]{\href{https://behavior-robot-suite.github.io/}{\textcolor{cardinal}{\textbf{\texttt{behavior-robot-suite.github.io}}}}}
\newcommand{\fullname}[0]{\mbox{\textsc{BEHAVIOR Robot Suite}}\xspace}
\newcommand{\acronym}[0]{\mbox{BRS}\xspace}
\newcommand{\interfacename}[0]{\mbox{JoyLo}\xspace}
\newcommand{\algoname}[0]{\mbox{WB-VIMA}\xspace}

\title{\textsc{BEHAVIOR Robot Suite:} Streamlining\\Real-World Whole-Body Manipulation\\for Everyday Household Activities}

\author{
  Yunfan Jiang, Ruohan Zhang, Josiah Wong, Chen Wang, Yanjie Ze,\\\textbf{Hang Yin, Cem Gokmen, Shuran Song, Jiajun Wu, Li Fei-Fei}\\
Stanford University\\
\webpage
}

\begin{document}
\maketitle

\vspace{-0.8cm}
\begin{abstract}
Real-world household tasks present significant challenges for mobile manipulation robots. An analysis of existing robotics benchmarks reveals that successful task performance hinges on three key whole-body control capabilities: bimanual coordination, stable and precise navigation, and extensive end-effector reachability. Achieving these capabilities requires careful hardware design, but the resulting system complexity further complicates visuomotor policy learning. To address these challenges, we introduce the \fullname (\acronym), a comprehensive framework for whole-body manipulation in diverse household tasks. Built on a bimanual, wheeled robot with a 4-DoF torso, \acronym integrates a cost-effective whole-body teleoperation interface for data collection and a novel algorithm for learning whole-body visuomotor policies. We evaluate \acronym on five challenging household tasks that not only emphasize the three core capabilities but also introduce additional complexities, such as long-range navigation, interaction with articulated and deformable objects, and manipulation in confined spaces. We believe that \acronym's integrated robotic embodiment, data collection interface, and learning framework mark a significant step toward enabling real-world whole-body manipulation for everyday household tasks.
\acronym is open-sourced at \webpage.
\end{abstract}


\begin{figure}[h]
\centering
\vspace{-0.2cm}
\includegraphics[width=\textwidth]{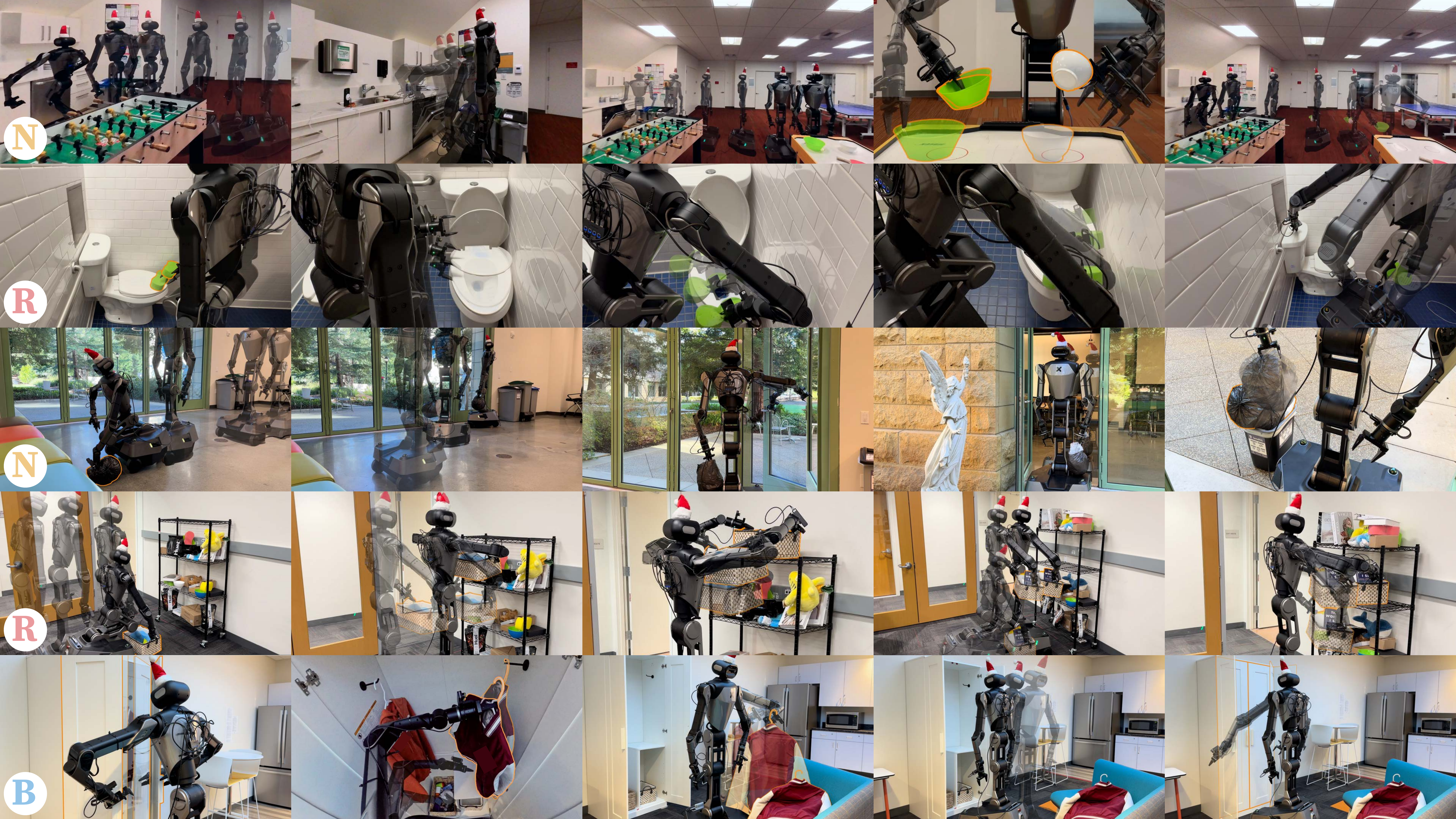}
\caption{\textbf{Everyday household activities enabled by \fullname (\acronym), showcasing its three core capabilities: \bimanual{bimanual} coordination (\bimanual{B}), stable and accurate \mobility{navigation} (\mobility{N}), and extensive end-effector \reachability{reachability} (\reachability{R}).}
}
\label{fig:pull}
\vspace{-0.7cm}
\end{figure}

\section{Introduction}
\label{sec:introduction}

Developing versatile and capable robots that can assist in everyday life remains a major challenge in human-centered robotics research  \citep{littman2022gathering,https://doi.org/10.1002/hbe2.117,10.1145/3328485,10.1145/3419764}, with increasing attention on daily household tasks~\citep{batra2020rearrangement,srivastava2021behavior,NEURIPS2021_021bbc7e,li2022behaviork,heo2023furniturebench,yenamandra2023homerobot,shukla2024maniskillhab,nasiriany2024robocasa}.
\emph{What key capabilities must a robot develop to achieve all these?} To investigate this question, we analyze activities from BEHAVIOR-1K~\citep{li2022behaviork}, a human-centered robotics benchmark encompassing 1,000 everyday household tasks, selected and defined by the general public, and instantiated in ecological and virtual environments.
Through this analysis, we identify three essential whole-body control capabilities for successfully performing these tasks: \bimanual{bimanual} coordination, stable and accurate \mobility{navigation}, and extensive end-effector \reachability{reachability}. 

\begin{table}[t]
  \vspace{-0.3cm}
    \centering
    \resizebox{\linewidth}{!}{
    \begin{threeparttable}
    \begin{tabular}{|c|c|c|ccccccccccccccc|}
        \rot{} & 
        \rot{} & 
        \rot{\textbf{\acronym (Ours)}} & 
        \rot{Mobile ALOHA~\citep{fu2024mobile}} & 
        \rot{TidyBot++~\citep{wu2024tidybot}} & 
        \rot{ACE~\citep{yang2024ace}} & 
        \rot{BiDex~\citep{shaw2024bimanual}} &
        \rot{HomeRobot~\citep{yenamandra2023homerobot}} & 
        \rot{TTT~\citep{bajracharya2023demonstrating}} & 
        \rot{TeleMoMa~\citep{dass2024telemoma}} &
        \rot{RoboCopilot~\citep{wu2025robocopilot0}} & 
        \rot{Open-TeleVision~\citep{cheng2024opentelevision}} & 
        \rot{TRILL~\citep{seo2023trill}} & 
        \rot{GR00T N1~\citep{nvidia2025gr00t}} &
        \rot{ALOHA~\citep{zhao2023learning}} & 
        \rot{GELLO~\citep{wu2023gello}} &
        \rot{HATO~\citep{lin2024learning}} & 
        \rot{FACTR~\citep{liu2025factr}}
        \\\cline{3-18}
        \multicolumn{2}{c}{} & 
        \multicolumn{9}{|c}{Mobile manipulation} & 
        \multicolumn{3}{|c}{Humanoid manipulation} & 
        \multicolumn{4}{|c|}{Stationary manipulation}
        \\\cline{2-18}
        \hline
        \rule{0pt}{12pt} & 
         \begin{tabular}{@{}c@{}}Simultaneous control of\\arms, torso, and mobile base\end{tabular} & 
         \textcolor{forestgreen}\faCheck & 
         \textcolor{red}\faTimes & 
         \textcolor{red}\faTimes & 
         \textcolor{red}\faTimes & 
         \textcolor{red}\faTimes &
         \textcolor{red}\faTimes & 
         \textcolor{red}\faTimes & 
         \textcolor{forestgreen}\faCheck & 
         \textcolor{forestgreen}\faCheck & 
         \textcolor{red}\faTimes &
         \textcolor{red}\faTimes &
         \textcolor{red}\faTimes & 
         \textcolor{red}\faTimes & 
         \textcolor{red}\faTimes &
         \textcolor{red}\faTimes &
         \textcolor{red}\faTimes
        \\
        \multirow{1}*{\rotatebox{90}{\textbf{Interface}\phantom{0}}}
        & 
        \begin{tabular}{@{}c@{}} Bimanual control \end{tabular} & 
        \textcolor{forestgreen}\faCheck & 
        \textcolor{forestgreen}\faCheck & 
        \textcolor{red}\faTimes &
        \textcolor{forestgreen}\faCheck &
        \textcolor{forestgreen}\faCheck & 
        \textcolor{red}\faTimes &
        \textcolor{red}\faTimes &
        \textcolor{forestgreen}\faCheck &
        \textcolor{forestgreen}\faCheck & 
        \textcolor{forestgreen}\faCheck & 
        \textcolor{forestgreen}\faCheck & 
        \textcolor{forestgreen}\faCheck & 
        \textcolor{forestgreen}\faCheck & 
        \textcolor{forestgreen}\faCheck & 
        \textcolor{forestgreen}\faCheck & 
        \textcolor{forestgreen}\faCheck
        \\
        & 
        \begin{tabular}{@{}c@{}}Torso control\end{tabular} &
        \textcolor{forestgreen}\faCheck & 
        \textcolor{red}\faTimes & 
        \textcolor{red}\faTimes & 
        \textcolor{red}\faTimes & 
        \textcolor{red}\faTimes & 
        \textcolor{red}\faTimes & 
        \textcolor{red}\faTimes & 
        \textcolor{forestgreen}\faCheck &
        \textcolor{forestgreen}\faCheck & 
        \textcolor{red}\faTimes & 
        \textcolor{red}\faTimes & 
        \textcolor{red}\faTimes & 
        \textcolor{red}\faTimes & 
        \textcolor{red}\faTimes &
        \textcolor{red}\faTimes &
        \textcolor{red}\faTimes
        \\

         & 
         \begin{tabular}{@{}c@{}}Single-operator mobile base control\end{tabular} & 
         \textcolor{forestgreen}\faCheck & 
         \textcolor{forestgreen}\faCheck & 
         \textcolor{forestgreen}\faCheck & 
         \textcolor{forestgreen}\faCheck & 
         \textcolor{red}\faTimes & 
         \textcolor{red}\faTimes & 
         \textcolor{red}\faTimes & 
         \textcolor{forestgreen}\faCheck & 
         \textcolor{forestgreen}\faCheck & 
         \textcolor{red}\faTimes &
         \textcolor{forestgreen}\faCheck &
         \textcolor{red}\faTimes &
         \textcolor{red}\faTimes &
         \textcolor{red}\faTimes &
         \textcolor{red}\faTimes &
         \textcolor{red}\faTimes
        \\
        & 
        \begin{tabular}{@{}c@{}}Untethered mobile base control\end{tabular} & 
        \textcolor{forestgreen}\faCheck & 
        \textcolor{red}\faTimes & 
        \textcolor{forestgreen}\faCheck &
        \textcolor{forestgreen}\faCheck & 
        \textcolor{forestgreen}\faCheck &
        \textcolor{red}\faTimes & 
        \textcolor{red}\faTimes & 
        \textcolor{forestgreen}\faCheck &
        \textcolor{forestgreen}\faCheck &
        \textcolor{red}\faTimes & 
        \textcolor{forestgreen}\faCheck & 
        \textcolor{red}\faTimes & 
        \textcolor{red}\faTimes & 
        \textcolor{red}\faTimes &
        \textcolor{red}\faTimes &
        \textcolor{red}\faTimes
        \\
        &
        \begin{tabular}{@{}c@{}}Haptic feedback\end{tabular} & 
        \textcolor{forestgreen}\faCheck & 
        \textcolor{red}\faTimes & 
        \textcolor{red}\faTimes &
        \textcolor{red}\faTimes &
        \textcolor{red}\faTimes &
        \textcolor{red}\faTimes & 
        \textcolor{red}\faTimes & 
        \textcolor{red}\faTimes &
        \textcolor{forestgreen}\faCheck &
        \textcolor{red}\faTimes &
        \textcolor{red}\faTimes &
        \textcolor{red}\faTimes &
        \textcolor{red}\faTimes &
        \textcolor{red}\faTimes &
        \textcolor{red}\faTimes &
        \textcolor{forestgreen}\faCheck
        \\        
        & 
        \begin{tabular}{@{}c@{}}Cost$^{(a)}$\end{tabular} & 
        \faDollar & 
        \faDollar\faDollar\faDollar & 
        \faDollar\faDollar & 
        \faDollar\faDollar &
        \faDollar &
        N.A. &
        N.A. & 
        \faDollar & 
        \faDollar\faDollar\faDollar & 
        \faDollar\faDollar\faDollar &
        \faDollar & 
        \faDollar\faDollar\faDollar &
        \faDollar\faDollar\faDollar &
        \faDollar &
        \faDollar &
        \faDollar\faDollar
        \\
        \cline{1-18}

        \rule{0pt}{12pt} 
        \multirow{1}*{\rotatebox{90}{\textbf{Capabilities}$^{(b)}$\phantom{0}}}
        & 
        \begin{tabular}{@{}c@{}}Omnidirectional navigation\end{tabular} & 
        \textcolor{forestgreen}\faCheck &
        \textcolor{red}\faTimes &
        \textcolor{forestgreen}\faCheck &
        \textcolor{red}\faTimes &
        \textcolor{forestgreen}\faCheck &
        \textcolor{red}\faTimes &
        \textcolor{forestgreen}\faCheck &
        \textcolor{forestgreen}\faCheck &
        \textcolor{forestgreen}\faCheck &
        \textcolor{red}\faTimes &
        \textcolor{red}\faTimes &
        \textcolor{red}\faTimes &
        \textcolor{red}\faTimes &
        \textcolor{red}\faTimes &
        \textcolor{red}\faTimes &
        \textcolor{red}\faTimes
        \\
        & 
        \begin{tabular}{@{}c@{}}Bimanual coordination\end{tabular} & 
        \textcolor{forestgreen}\faCheck &
        \textcolor{forestgreen}\faCheck &
        \textcolor{red}\faTimes &
        \textcolor{forestgreen}\faCheck &
        \textcolor{forestgreen}\faCheck &
        \textcolor{red}\faTimes &
        \textcolor{forestgreen}\faCheck &
        \textcolor{forestgreen}\faCheck &
        \textcolor{forestgreen}\faCheck &
        \textcolor{forestgreen}\faCheck &
        \textcolor{forestgreen}\faCheck &
        \textcolor{forestgreen}\faCheck &
        \textcolor{forestgreen}\faCheck &
        \textcolor{red}\faTimes &
        \textcolor{forestgreen}\faCheck &
        \textcolor{forestgreen}\faCheck
        \\ 
        &  
        \begin{tabular}{@{}c@{}}Ground-level reach\end{tabular} & 
        \textcolor{forestgreen}\faCheck &
        \textcolor{red}\faTimes &
        \textcolor{forestgreen}\faCheck &
        \textcolor{forestgreen}\faCheck &
        \textcolor{red}\faTimes &
        \textcolor{red}\faTimes &
        \textcolor{forestgreen}\faCheck &
        \textcolor{forestgreen}\faCheck &
        \textcolor{forestgreen}\faCheck &
        \textcolor{red}\faTimes &
        \textcolor{red}\faTimes &
        \textcolor{red}\faTimes &
        \textcolor{red}\faTimes &
        \textcolor{red}\faTimes &
        \textcolor{red}\faTimes &
        \textcolor{red}\faTimes
        \\ 
        & 
        \begin{tabular}{@{}c@{}}Comfortable overhead reach$^{(c)}$\end{tabular} & 
        \textcolor{forestgreen}\faCheck &
        \textcolor{forestgreen}\faCheck &
        \textcolor{red}\faTimes &
        \textcolor{red}\faTimes &
        \textcolor{red}\faTimes &
        \textcolor{red}\faTimes &
        \textcolor{forestgreen}\faCheck &
        \textcolor{red}\faTimes &
        \textcolor{red}\faTimes &
        \textcolor{red}\faTimes &
        \textcolor{red}\faTimes &
        \textcolor{red}\faTimes &
        \textcolor{red}\faTimes &
        \textcolor{red}\faTimes &
        \textcolor{red}\faTimes &
        \textcolor{red}\faTimes
        \\ 
        & 
        \begin{tabular}{@{}c@{}}Operation in confined spaces\end{tabular} & 
        \textcolor{forestgreen}\faCheck & 
        \textcolor{red}\faTimes &
        \textcolor{forestgreen}\faCheck &
        \textcolor{red}\faTimes &
        \textcolor{red}\faTimes &
        \textcolor{red}\faTimes &
        \textcolor{red}\faTimes &
        \textcolor{red}\faTimes &
        \textcolor{red}\faTimes &
        \textcolor{red}\faTimes &
        \textcolor{red}\faTimes &
        \textcolor{red}\faTimes &
        \textcolor{red}\faTimes &
        \textcolor{red}\faTimes &
        \textcolor{red}\faTimes &
        \textcolor{red}\faTimes
        \\ 
        & 
         \begin{tabular}{@{}c@{}}Coordinated whole-body manipulation\\involving hip, waist, and mobile base\end{tabular} & 
        \textcolor{forestgreen}\faCheck &
        \textcolor{red}\faTimes &
        \textcolor{red}\faTimes &
        \textcolor{red}\faTimes &
        \textcolor{red}\faTimes &
        \textcolor{red}\faTimes &
        \textcolor{red}\faTimes &
        \textcolor{red}\faTimes &
        \textcolor{red}\faTimes &
        \textcolor{red}\faTimes &
        \textcolor{red}\faTimes &
        \textcolor{red}\faTimes &
        \textcolor{red}\faTimes &
        \textcolor{red}\faTimes &
        \textcolor{red}\faTimes &
        \textcolor{red}\faTimes
        \\\cline{1-18}

        \rule{0pt}{12pt} 
        & 
        \begin{tabular}{@{}c@{}}Learning-based method\end{tabular} & 
        \textcolor{forestgreen}\faCheck &
        \textcolor{forestgreen}\faCheck &
        \textcolor{forestgreen}\faCheck &
        \textcolor{forestgreen}\faCheck &
        \textcolor{forestgreen}\faCheck &
        \textcolor{forestgreen}\faCheck &
        \textcolor{red}\faTimes &
        \textcolor{forestgreen}\faCheck &
        \textcolor{forestgreen}\faCheck &
        \textcolor{forestgreen}\faCheck &
        \textcolor{forestgreen}\faCheck &
        \textcolor{forestgreen}\faCheck &
        \textcolor{forestgreen}\faCheck &
        N.A. &
        \textcolor{forestgreen}\faCheck &
        \textcolor{forestgreen}\faCheck
        \\
        \multirow{1}*{\rotatebox{90}{\textbf{Algorithm}\phantom{0}}}
        & 
        \begin{tabular}{@{}c@{}}Novel algorithm\end{tabular} & 
        \textcolor{forestgreen}\faCheck &
        \textcolor{red}\faTimes &
        \textcolor{red}\faTimes &
        \textcolor{red}\faTimes &
        \textcolor{red}\faTimes &
        \textcolor{red}\faTimes &
        \textcolor{red}\faTimes &
        \textcolor{red}\faTimes &
        \textcolor{red}\faTimes &
        \textcolor{red}\faTimes &
        \textcolor{red}\faTimes &
        \textcolor{forestgreen}\faCheck &
        \textcolor{forestgreen}\faCheck &
        N.A. &
        \textcolor{red}\faTimes &
        \textcolor{forestgreen}\faCheck
        \\ 
        &
        \begin{tabular}{@{}c@{}}Autoregressive whole-\\body action prediction\end{tabular} & 
        \textcolor{forestgreen}\faCheck &
        \textcolor{red}\faTimes &
        \textcolor{red}\faTimes &
        \textcolor{red}\faTimes &
        \textcolor{red}\faTimes &
        \textcolor{red}\faTimes &
        \textcolor{red}\faTimes &
        \textcolor{red}\faTimes &
        \textcolor{red}\faTimes &
        \textcolor{red}\faTimes &
        \textcolor{red}\faTimes &
        \textcolor{red}\faTimes &
        \textcolor{red}\faTimes &
        N.A. &
        \textcolor{red}\faTimes &
        \textcolor{red}\faTimes
        \\ 
        & 
        \begin{tabular}{@{}c@{}}Sensory observation modality\end{tabular} & 
        \begin{tabular}{@{}c@{}}\textbf{Colored}\\\textbf{point cloud}\end{tabular} &
        RGB &
        RGB &
        RGB &
        RGB &
        \begin{tabular}{@{}c@{}}Depth +\\semantic seg.\end{tabular} &
        RGB-D &
        RGB-D &
        RGB &
        RGB &
        RGB &
        RGB &
        RGB &
        N.A. &
        \begin{tabular}{@{}c@{}}RGB-D +\\tactile\end{tabular} &
        RGB
        \\ 
        & 
        \begin{tabular}{@{}c@{}}Policy model backbone$^{(d)}$\end{tabular} &
        \textbf{XF} &
        XF &
        UNet &
        UNet &
        XF &
        RNN &
        N.A. &
        MLP/RNN &
        UNet &
        XF &
        RNN &
        XF &
        XF &
        N.A. &
        UNet &
        XF
        \\\cline{1-18}

        \rot{} 
        \rule{0pt}{15pt} & 
        \multicolumn{1}{|c|}{\begin{tabular}{@{}c@{}}Open-source everything$^{(e)}$\end{tabular}} &
        \textcolor{forestgreen}\faCheck &
        \textcolor{forestgreen}\faCheck &
        \textcolor{forestgreen}\faCheck &
        \begin{tabular}{@{}c@{}}Hardware\\+ teleop.\end{tabular} &
        Teleop. &
        \textcolor{forestgreen}\faCheck &
        \textcolor{red}\faTimes &
        Teleop. &
        \textcolor{red}\faTimes &
        \textcolor{forestgreen}\faCheck &
        \textcolor{forestgreen}\faCheck &
        \begin{tabular}{@{}c@{}}Weights\\+ finetuning\end{tabular} &
        \textcolor{forestgreen}\faCheck &
        \textcolor{forestgreen}\faCheck &
        \textcolor{forestgreen}\faCheck &
        \textcolor{forestgreen}\faCheck
        \\\cline{2-18}
    \end{tabular}
    \begin{tablenotes}
    \item[{(a)}]{Interface hardware cost. \faDollar: \SI{0}[\$] - \SI{500}[\$]; \faDollar\faDollar: \SI{500}[\$] - \SI{1000}[\$]; \faDollar\faDollar\faDollar: \SI{1000}[\$]+.}
    \item[{(b)}]{We consider robot capabilities that are demonstrated by \textbf{learned autonomous policies}.}
    \item[{(c)}]{Following \citet{panero2014human}, we use \SI{182.9}{\centi\meter} (\SI{72}{in}) as the maximum height for comfortable overhead reach.}
    \item[{(d)}]{Neural network architecture of the policy backbone. ``XF'' stands for Transformer.}
    \item[{(e)}]{Everything includes the interface hardware, teleoperation code, algorithm code, and documentation.}
    \end{tablenotes}
    \end{threeparttable}
  }
  \vspace{0.1cm}
\caption{\textbf{Comparison of recent real-robot frameworks.} \acronym is comprehensive, integrating a unique whole-body control interface \interfacename and a novel algorithm \algoname for learning whole-body visuomotor policies, demonstrating several unprecedented robotic capabilities.}
  \label{table:comparison}
  \vspace{-0.4cm}
\end{table}

\begin{wrapfigure}[19]{r}{0.4\textwidth}
\vspace{-0.3cm}
    \centering
    \includegraphics[width=0.4\textwidth]{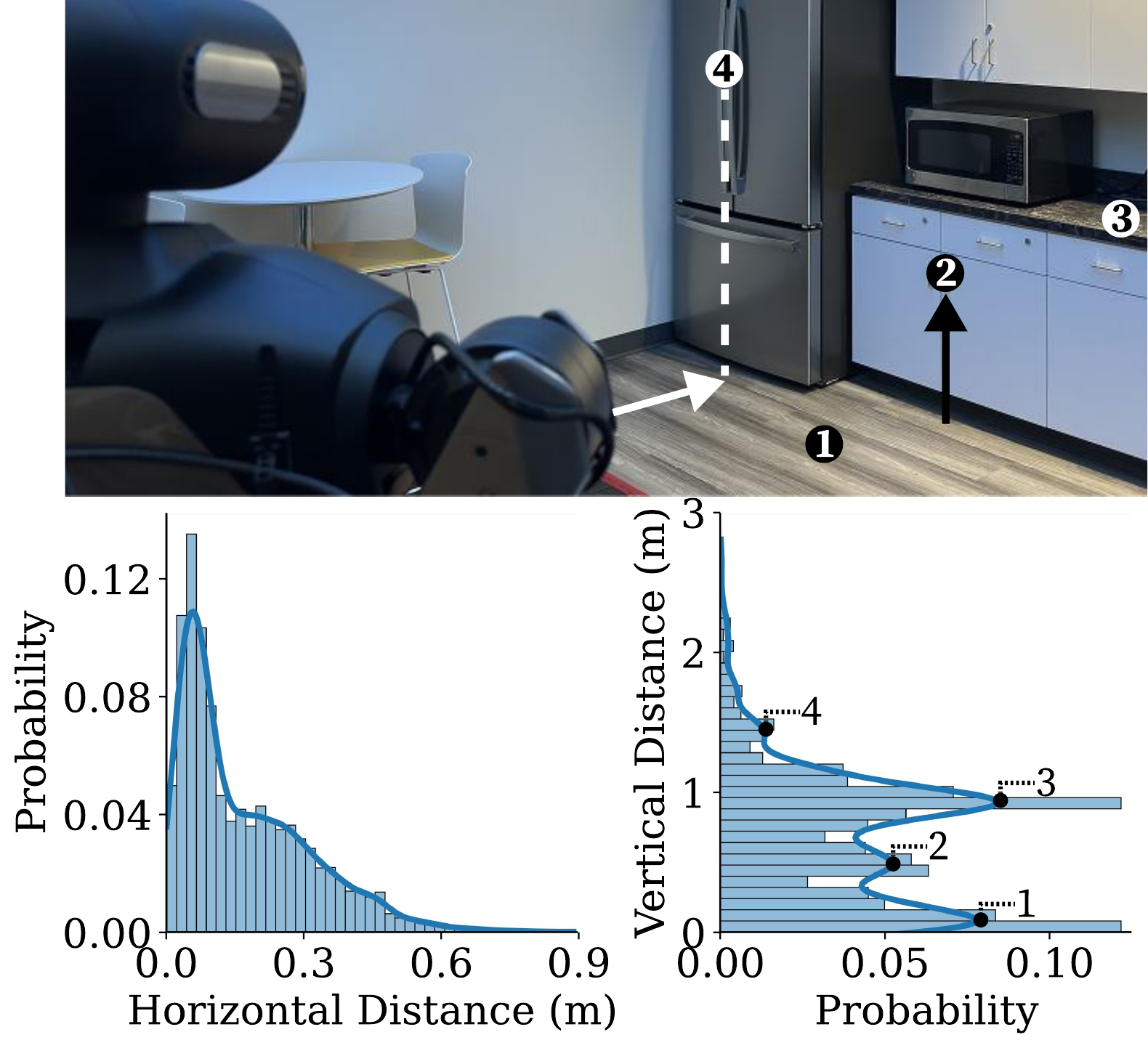}
    \vspace{-0.6cm}
    \caption{\textbf{Ecological distributions of task-relevant objects in daily household activities.} Multiple distinct modes appear in the vertical distance distribution, located at \SI{0.09}{\meter}, \SI{0.49}{\meter}, \SI{0.94}{\meter}, and \SI{1.43}{\meter}, representing heights at which objects are typically found.} 
    \label{fig:b1k_object_spatial_dist}
\end{wrapfigure}

Tasks such as lifting large, heavy objects require \bimanual{bimanual manipulation}~\citep{SMITH20121340,doi:10.1126/science.aat8414}, whereas retrieving objects throughout a house depends on stable and precise \mobility{navigation}~\citep{982903,KRUSE20131726,xiao22motion}. Opening a door while carrying groceries demands the coordination of both capabilities~\citep{895316,7363442,https://doi.org/10.1002/rob.21685}.
In addition, everyday objects are distributed across diverse locations and heights, requiring robots to adapt their \reachability{reach} accordingly. To illustrate this, we analyze the spatial distribution of task-relevant household objects in everyday household tasks and scenes (Fig.~\ref{fig:b1k_object_spatial_dist}).
Notably, the multi-modal distribution of vertical distances highlights the necessity of extensive end-effector reachability, enabling a robot to interact with objects across a wide range of spatial configurations.

How, then, can a robot effectively achieve these capabilities? Carefully designed robotic hardware incorporating dual arms, a mobile base, and a flexible torso is essential to enable whole-body manipulation~\citep{bajracharya2023demonstrating}. However, such designs introduce significant challenges for policy learning methods, particularly in scaling data collection~\citep{10611477,walke2023bridgedata,khazatsky2024droid} and accurately modeling coordinated whole-body actions. Current systems struggle to address these challenges comprehensively~\citep{bajracharya2023demonstrating,dass2024telemoma,shafiullah2023bringing,hu2023causal,jiang2024harmon,uppal2024spin,fu2024mobile,xiong2024adaptive}, highlighting the need for more suitable hardware for household tasks, more efficient data collection tools, and improved models for whole-body control.

We introduce the \fullname (\acronym), a comprehensive framework for learning whole-body manipulation to tackle diverse real-world household tasks (Fig.~\ref{fig:pull}). \acronym addresses both hardware and learning challenges through two key innovations (Table~\ref{table:comparison}). The first is \interfacename, a low-cost, whole-body teleoperation interface designed for general applicability, with a concrete implementation on a wheeled dual-arm manipulator with a flexible torso. The second is the Whole-Body VIsuoMotor Attention (\algoname) policy, a novel learning algorithm that effectively models coordinated whole-body actions.

We evaluate \acronym on five challenging real-world household tasks in unmodified human living environments.
The learned \algoname policies demonstrate strong performance,
achieving an average success rate of 88\% in short-horizon sub-tasks, and a peak success rate of 93\% in long-horizon full tasks.
We believe that BRS's integrated robotic embodiment, data collection interface, and learning framework mark a significant step toward real-world whole-body manipulation for everyday household tasks.
\acronym is open-sourced at \webpage.

\begin{figure}[t]
    \centering
    \includegraphics[width=\textwidth]{figs/system_fig-fig.pdf}
    \caption{\textbf{\acronym hardware system.} \textbf{Left:} The R1 robot with two 6-DoF arms and a 4-DoF torso mounted on an omnidirectional mobile base. \textbf{Right:} The \interfacename system, consisting of compact, off-the-shelf Nintendo Joy-Con controllers mounted at the ends of two kinematic-twin arms. Joy-Con serves as the interface for controlling the grippers, torso, and mobile base.}
    \label{fig:hardware_system}
    \vspace{-0.5cm}
\end{figure}

\section{JoyLo: Joy-Con on Low-Cost Kinematic-Twin Arms}
\label{sec:system}
\label{sec:joylo}
To enable seamless teleoperation of mobile manipulators with a high degree of freedoms (DoFs) and facilitate data collection for policy learning, we introduce \textbf{JoyLo}, a cost-effective whole-body teleoperation interface.
As illustrated in Fig.~\ref{fig:hardware_system}, we implement \interfacename on the Galaxea R1 robot, a wheeled dual-arm manipulator with a 4-DoF torso (Appendix~\ref{appendix:sec:hardware_details}), following design objectives detailed as follow.
While we provide one specific instantiation of \interfacename, its design principles are general and can be adapted to similar mobile manipulators.

\para{Efficient Whole-Body Control}
Whole-body robot teleoperation methods vary widely in accuracy, efficiency, applicability, and user experience. 
At one extreme, kinesthetic teaching enables precise physical guidance~\citep{Kormushev01012011,annurev:/content/journals/10.1146/annurev-control-100819-063206,10.5898/JHRI.2.1.Wrede,hagenow2024versatile}, but is slow and not easily scalable. At the other extreme, motion retargeting techniques~\citep{10.5555/1838206.1838419,stanton2012teleop,9555769,dass2024telemoma,Antotsiou_2018_ECCV_Workshops,8794277,9197117,9197124,9849105,sivakumar2022robotic,qin2023anyteleop} remove physical interaction but face embodiment mismatches and limited platform applicability.
To balance intuitiveness, ease of use, and precision for manipulation tasks,
we propose a puppeteering-based approach using kinematic-twin arms equipped with thumbsticks for torso and mobile base control.
Specifically, we utilize off-the-shelf Nintendo Joy-Con controllers due to their compact size, integrated thumbsticks, and multiple functional buttons, which enable rich, customizable functionality.
As illustrated in Fig.~\ref{fig:hardware_system}, the left thumbstick controls mobile base velocity; the right thumbstick adjusts waist and hips; arrow keys change torso height; triggers operate the grippers.
With \interfacename, users can simultaneously control arm movements, gripper operations, upper-body motions, and mobile base navigation, enabling efficient whole-body control that is accurate, user-friendly, and scalable.
Additionally, the kinematic constraints imposed by the leader arms prevent the operator from generating infeasible or undeployable actions, ensuring smooth and reliable demonstrations.

\para{Rich User Feedback}\label{sec:bilateral_joylo}
\interfacename enhances teleoperation by providing haptic feedback through bilateral teleoperation~\citep{88057,258054} without extra force sensors~\citep{https://doi.org/10.1002/rob.21960,LI2016188}.
The \interfacename arms, kinematically coupled with the robot arms, act as leaders issuing commands while being regularized by the robot’s joint positions.
Let $\mathbf{q}_\text{JoyLo}$ and $\mathbf{q}_\text{robot}$ be their respective joint positions; the torques $\tau$ applied to the \interfacename arms are $\tau = \mathbf{K_p}\left( \mathbf{q}_\text{robot} - \mathbf{q}_\text{JoyLo} \right) + \mathbf{K_d}(\dot{\mathbf{q}}_\text{robot} - \dot{\mathbf{q}}_\text{JoyLo}) - \mathbf{K}$,
where $\dot{\mathbf{q}}$ denotes joint velocities, and $\mathbf{K_p}$, $\mathbf{K_d}$, and $\mathbf{K}$ are proportional, derivative, and damping gains.
This feedback discourages abrupt user motions and provides proportional resistance when the robot experiences contact.

\para{Low Cost and Easy Accessibility}
\interfacename is built from 3D-printed links, low-cost Dynamixel motors, and Joy-Con controllers, totaling under \$500.
Additionally, its modular design ensures that all components are replaceable, minimizing downtime and eliminating unnecessary repair costs.
\acronym provides an intuitive, real-time controller with Python interfaces for efficient operation.

\section{WB-VIMA: Whole-Body VIsuoMotor Attention Policy}
\label{sec:method}
\begin{figure}[t]
    \centering
    \includegraphics[width=\textwidth]{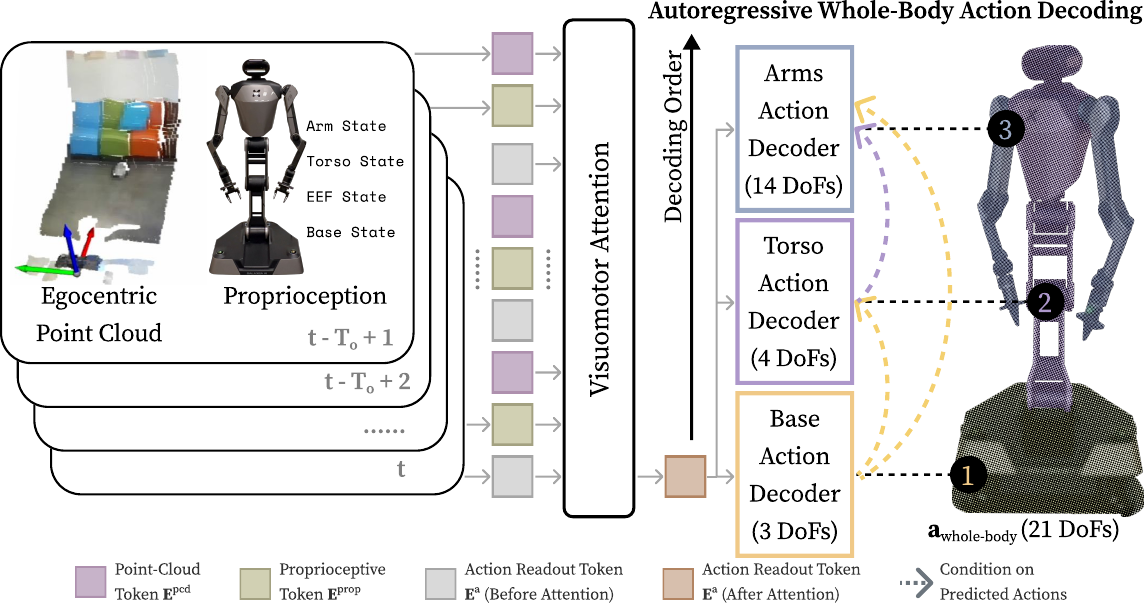}
    \caption{\textbf{\algoname architecture.} It autoregressively decodes whole-body actions by leveraging the hierarchical  interdependencies within the embodiment space, and dynamically aggregates multi-modal observations using self-attention.}
    \label{fig:model_arch}
    \vspace{-0.5cm}
\end{figure}

This section introduces \textbf{\algoname}, a transformer-based model~\citep{vaswani2017attention,jiang2022vima} designed to learn coordinated whole-body actions for mobile manipulation tasks.
Trained on data collected through \interfacename, it autoregressively decodes whole-body actions across the embodiment space and dynamically aggregates multi-modal observations using self-attention (Fig.~\ref{fig:model_arch}).

\para{Autoregressive Whole-Body Action Decoding}\label{sec:method:wb_vima}

In mobile manipulators with multiple articulated components, small mobile base or torso errors can cause large end-effector deviations.
For example, a \SI{0.17}{\radian} (\SI{10}{\degree}) knee movement in the R1 robot’s neutral pose (Fig.~\ref{fig:hardware_system}) can shift the end-effector by up to \SI{0.14}{\meter} due to error amplification along the kinematic chain, highlighting the need for precise coordination in whole-body mobile manipulation.
To address this issue, we leverage the inherent hierarchy in the robot’s embodiment. Specifically, conditioning upper-body action predictions on the predicted lower-body actions enables the policy to better model coordinated whole-body movements. This approach ensures that downstream joints account for upstream motion, reducing error propagation.
The whole-body action decoding follows an autoregressive structure:
At timestep $t$, the mobile base trajectory $\mathbf{a}_\text{base} \in \mathbb{R}^{T_a \times 3}$ is first predicted using the action readout token $\mathbf{E}^a$ (encoded from observations, detailed later).
$\mathbf{a}_\text{base}$ and $\mathbf{E}^a$ are then used to predict the torso trajectory $\mathbf{a}_\text{torso} \in \mathbb{R}^{T_a \times 4}$.
Finally, $\mathbf{a}_\text{base}$,  $\mathbf{a}_\text{torso}$, and $\mathbf{E}^a$ together predict the arms and grippers’ trajectory $\mathbf{a}_\text{arms} \in \mathbb{R}^{T_a \times 14}$.
\algoname jointly learns three independent denoising diffusion networks~\citep{wang2022diffusion,chi2023diffusion,wang2024onestep} for the mobile base, torso, and arms, denoted $\epsilon_{\text{base}}$, $\epsilon_{\text{torso}}$, and $\epsilon_{\text{arms}}$.
Whole-body actions $\mathbf{a}_\text{whole-body} \in \mathbb{R}^{T_a \times 21}$ are autoregressively decoded through iterative denoising:
\begin{equation}
    \begin{split}
        \mathbf{a}^{k - 1}_\text{base} &\sim \mathcal{N}\left(\mu_k\left(\mathbf{a}^k_\text{base}, \epsilon_\text{base} \left(\mathbf{a}^k_\text{base} \vert \mathbf{E}^a, k\right) \right), \sigma_k^2I\right),\\
        \mathbf{a}^{k - 1}_\text{torso} &\sim \mathcal{N}\left(\mu_k\left(\mathbf{a}^k_\text{torso}, \epsilon_\text{torso} \left(\mathbf{a}^k_\text{torso} \vert \mathbf{a}^0_\text{base}, \mathbf{E}^a, k\right) \right), \sigma_k^2I\right),\\
        \mathbf{a}^{k - 1}_\text{arms} &\sim \mathcal{N}\left(\mu_k\left(\mathbf{a}^k_\text{arms}, \epsilon_\text{arms} \left(\mathbf{a}^k_\text{arms} \vert \mathbf{a}^0_\text{torso}, \mathbf{a}^0_\text{base}, \mathbf{E}^a, k\right) \right), \sigma_k^2I\right).
    \end{split}
\end{equation}
To achieve efficient inference for high-frequency control, only action readout tokens are used for whole-body decoding via diffusion, allowing lightweight UNet-based~\citep{ronneberger2015unet} action heads with a heavier transformer backbone for observation encoding. This balances expressivity and latency.

\para{Multi-Modal Observation Attention}
\label{sec:multi-modal-obs-attention}
Observations from multiple modalities are crucial for autonomous robots in complex environments.
In \algoname, egocentric colored point clouds and robot proprioception (joint positions and mobile base velocities) are fused via a visuomotor attention network, avoiding overfitting to any single source of information.
Concretely, a PointNet~\citep{qi2016pointnet} encodes the point cloud into a point-cloud token $\mathbf{E}^\text{pcd}$, and an MLP encodes proprioception into a proprioceptive token $\mathbf{E}^\text{prop}$.
Tokens from current and past $T_o$ steps, along with action readout tokens $\mathbf{E}^\text{a}$, form a visuomotor sequence: $\mathbf{S} = [\mathbf{E}^\text{pcd}_{t - T_o + 1}, \mathbf{E}^\text{prop}_{t - T_o + 1}, \mathbf{E}^\text{a}_{t - T_o + 1},\ldots, \mathbf{E}^\text{pcd}_{t}, \mathbf{E}^\text{prop}_{t}, \mathbf{E}^\text{a}_{t}  ] \in \mathbb{R}^{3T_o \times E}$.
$\mathbf{S}$ is then processed through causal self-attention, ensuring action tokens attend only to earlier observations. The final action readout token $\mathbf{E}^a_t$ is used for autoregressive whole-body decoding.

\para{Training and Deployment}
Following \citet{NEURIPS2020_4c5bcfec}, \algoname is trained to predict added noise, minimizing $\mathcal{L} = MSE(\epsilon^k, \epsilon_\theta(\cdot \vert k))$ for each action decoder, with the total loss aggregated across all three action decoders. Here, $\epsilon^k$ and $\epsilon_\theta$ represent the ground-truth and predicted noise.
Deployment uses NVIDIA RTX 4090 GPUs with \SI{0.02}{\second} effective latency.
Data is collected at \SI{10}{\hertz} with the robot controller running at \SI{100}{\hertz}.
A new policy action is issued every \SI{0.1}{\second} and repeated 10 times.

\section{Experiments}
\label{sec:experiments}

We conduct experiments to answer the following questions. \q{1}:What household tasks are enabled by \acronym, and how does \algoname compare to baselines? \q{2}:How different components contribute to \algoname's effectiveness? \q{3}:{How does \interfacename compare to other interfaces in efficiency and policy learning suitability?} \q{4}:{What other insights can be drawn about the system's capabilities?}

\para{Experiment Settings} We evaluate \acronym on five real-world household tasks (see Fig.~\ref{fig:pull} and Appendix~\ref{appendix:sec:task_definition} for details), inspired by the everyday activities defined in BEHAVIOR-1K~\citep{li2022behaviork}.
We collect \textbf{100}, \textbf{103}, \textbf{98}, \textbf{138}, and \textbf{122} trajectories using \interfacename for these long-horizon tasks, each ranging from \SI{60}{\second} to \SI{210}{\second}.
Each task is segmented into multiple sub-tasks (\textbf{``ST''}).
During evaluation, if a sub-task fails, we reset to the start of the \emph{next} sub-task and \emph{continue} evaluation.
We also report the end-to-end success rates for entire tasks (\textbf{``ET''}).
Baselines include \textbf{DP3}~\citep{ze20243d}, \textbf{RGB-DP}~\citep{chi2023diffusion}, and \textbf{ACT}~\citep{zhao2023learning}.
We additionally report human teleoperation success and policy safety violations, defined as robot collisions or motor power losses due to excessive force.
Each policy is evaluated 15 times with randomized \textbf{robot starting position}, \textbf{target object placement}, \textbf{target object instance}, and \textbf{distractors}. Each task covers at least two types of randomization.
Task videos are available at \webpage{}.

\para{\acronym enables various household activities, on which \algoname consistently outperforms baseline methods (\q{1}).}
As shown in Fig.~\ref{fig:main_exp_result}, \algoname achieves an average sub-task success rate of 88\%, and average and peak entire-task success rates of 58\% and 93\%.
On contact-rich sub-tasks involving articulated objects, where human operators often struggle with uncoordinated whole-body motions---such as opening the toilet cover (ST-2) in ``clean the toilet'' and opening the wardrobe (ST-1) in ``lay clothes out''---\algoname even outperforms human teleoperation, suggesting that training on successful demonstrations enables it to learn precise, coordinated maneuvers for reliably completing such tasks.
Moreover, \algoname shows an emergent capability for completing long-horizon, multi-stage tasks, enabled by the synergy between its multi-modal observation attention---extracting salient, task-relevant features---and autoregressive whole-body action decoding---generating coherent actions that rarely lead to out-of-distribution states.
Finally, \algoname maintains a near-zero safety violation rate, which we attribute to its use of colored point-cloud observations that provide explicit 3D perception and semantic understanding, ensuring coordinated actions that inherently respect safety constraints.

\begin{figure}[t]
    \centering
    \includegraphics[width=\textwidth]{figs/main_exp_results_generalization_settings-fig.pdf}
    \caption{\textbf{Evaluation results for five household tasks.} \textbf{Left:} Initial randomization. \textbf{Middle:} Success rates over 15 runs (``ET'' = entire task, ``ST'' = sub-task). \textbf{Right:} Number of safety violations.}
    \label{fig:main_exp_result}
    \vspace{-0.5cm}
\end{figure}

For end-to-end task success, \algoname achieves 13$\times$ and 21$\times$ higher success rates than DP3 and RGB-DP, respectively.
For average sub-task performance, it outperforms them by 1.6$\times$ and 3.4$\times$. ACT fails to complete any full tasks and rarely succeeds in sub-tasks.
These baselines struggle because they directly predict flattened 21-DoF actions, ignoring hierarchical dependencies within the action space. As a result, modeling errors~\citep{ross2010reduction} in mobile base or torso predictions cannot be corrected by arm actions, leading to amplified end-effector drift, pushing the robot into out-of-distribution states, and eventually resulting in task failures.
Uncoordinated whole-body actions also increase safety violations (Fig.~\ref{fig:main_exp_result}), such as DP3 colliding with tables, RGB-DP losing arm power from excessive force, and ACT hitting doorframes during trash disposal.
We also observe that \algoname and DP3 outperform RGB-DP and ACT, underscoring the importance of explicit 3D perception in complex environments. Egocentric point clouds provide unified spatial understanding critical for accurate mobile base navigation. While both \algoname and DP3 leverage point clouds, only \algoname incorporates task semantic information through color, whereas DP3 often overfits to proprioception, stitching actions based purely on joint positions without regard to the environment.

\begin{wrapfigure}[16]{r}{0.4\textwidth}
    \vspace{-0.2cm}
    \centering
    \includegraphics[width=0.4\textwidth]{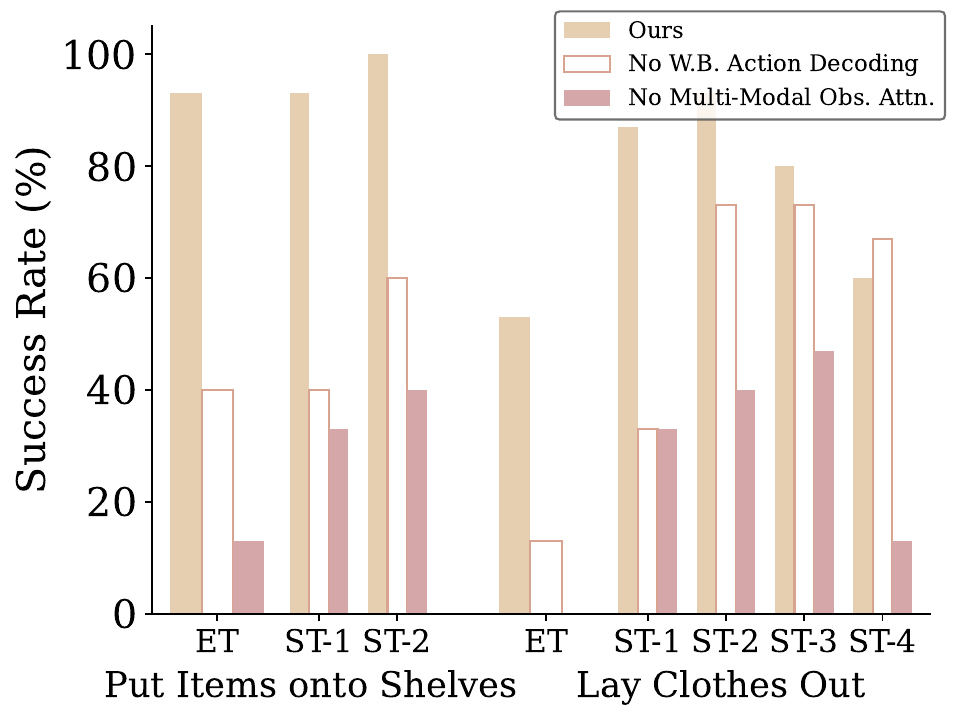}
    \resizebox{0.4\textwidth}{!}{
\begin{tabular}{@{}rccc@{}}
\toprule
\multicolumn{1}{c}{\textit{\begin{tabular}[c]{@{}c@{}}\# Safety\\ Violations\end{tabular}}} & Ours & \begin{tabular}[c]{@{}c@{}}No W.B.\\ Action Decoding\end{tabular} & \begin{tabular}[c]{@{}c@{}}No Multi-\\ Modal Obs. Attn.\end{tabular} \\ \midrule
\textbf{\begin{tabular}[c]{@{}r@{}}Put Items\\ onto Shelves\end{tabular}}                   & 0    & 0                                                                 & 0                                                                    \\
\textbf{Lay Clothes Out}                                                                    & 0    & 0                                                                 & 4                                                                    \\ \bottomrule
\end{tabular}
}
    \vspace{-0.2cm}
    \caption{\textbf{Real-world ablation results for ``put items onto shelves'' and ``lay clothes out.''}}
    \label{fig:ablation_results}
\end{wrapfigure}

\begin{wrapfigure}[25]{r}{0.4\textwidth}
    \vspace{-0.5cm}
    \centering
    \includegraphics[width=0.4\textwidth]{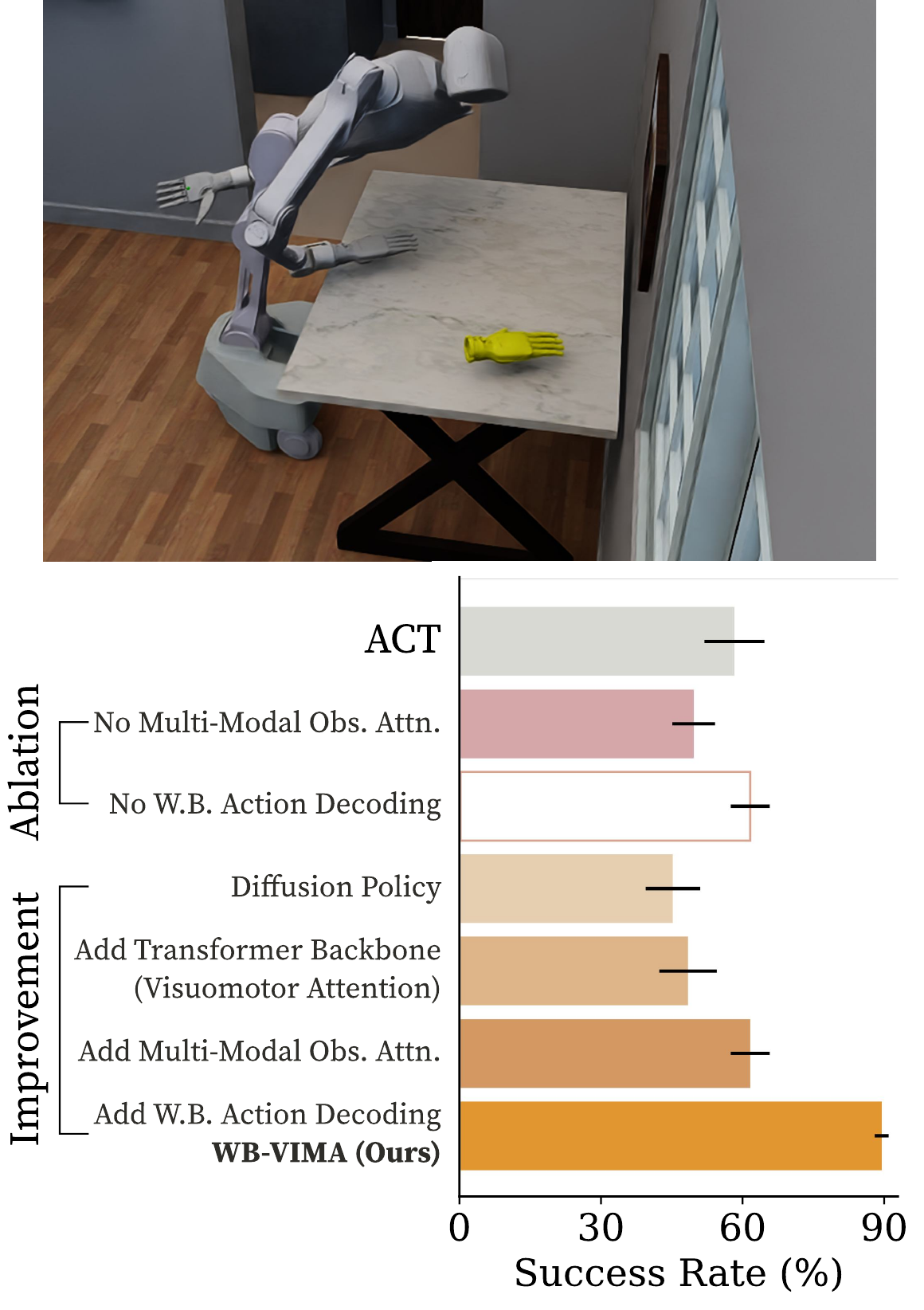}
    \vspace{-0.6cm}
    \caption{\textbf{Simulation ablation results for ``wiping table.''} The robot must wipe toward the goal using whole-body motions while maintaining continuous hand contact. Results are averaged over \textbf{five} runs with \textbf{100} rollouts each; error bars indicate standard deviation.} 
    \label{fig:sim_ablation}
\end{wrapfigure}

\para{Synergistic whole-body action prediction and multi-modal feature extraction are key to WB-VIMA’s strong performance (\q{2}).} Can models based solely on explicit 3D perception match WB-VIMA’s performance? Ablation studies show they cannot.
We evaluate two \algoname variants: one without \textbf{autoregressive whole-body action decoding} and one without \textbf{multi-modal observation attention}.
As shown in Fig.~\ref{fig:ablation_results}, removing either significantly degrades performance.
Tasks like ``put items onto shelves'' and ``open wardrobe'' (ST-1) in ``lay clothes out'' critically depend on coordinated whole-body actions; removing autoregressive action decoding leads to up to a 53\% performance drop.
Removing multi-modal attention reduces performance across all tasks, causing the model to ignore visual inputs and overfit to proprioception. Four collisions are also observed due to poor visual awareness.
The same conclusions hold in a simulated table wiping task (Fig.~\ref{fig:sim_ablation}). Furthermore, starting from a vanilla diffusion policy, we provide a roadmap improving the model success by progressively adding components: multi-modal observation attention improves by 27\% and surpasses ACT; adding autoregressive whole-body action decoding further boosts success by 45\%, culminating in WB-VIMA’s strong final performance.

\para{\interfacename is an efficient, user-friendly interface that provides high-quality data for policy learning (\q{3}).}\label{sec:exp:user_study}
We conducted a user study with 10 participants to evaluate \interfacename against two IK-based interfaces: \textbf{VR controllers}~\citep{dass2024telemoma} and \textbf{Apple Vision Pro}~\citep{cheng2024opentelevision,park2024avp}.
The study was performed in the OmniGibson simulator~\citep{li2022behaviork} on the ``clean house after a wild party'' task, with randomized interface exposure to eliminate bias.
We measured \emph{success rate}, \emph{completion time}, \emph{replay success rate}, and \emph{singularity ratio} across entire tasks and sub-tasks.
Replay success measures the open-loop execution of collected robot trajectories, where higher values indicate higher-quality, verified data that allows imitation learning policies to better model trajectories~\citep{mandlekar2021matters,fang2023airexo,shaw2024bimanual,yang2024ace,chen2024arcap}.
Further setup details are provided in Appendix~\ref{appendix:sec:user_study_details}.

\begin{wrapfigure}[13]{r}{0.5\textwidth}
    \vspace{-1cm}
    \centering
    \includegraphics[width=0.5\textwidth]{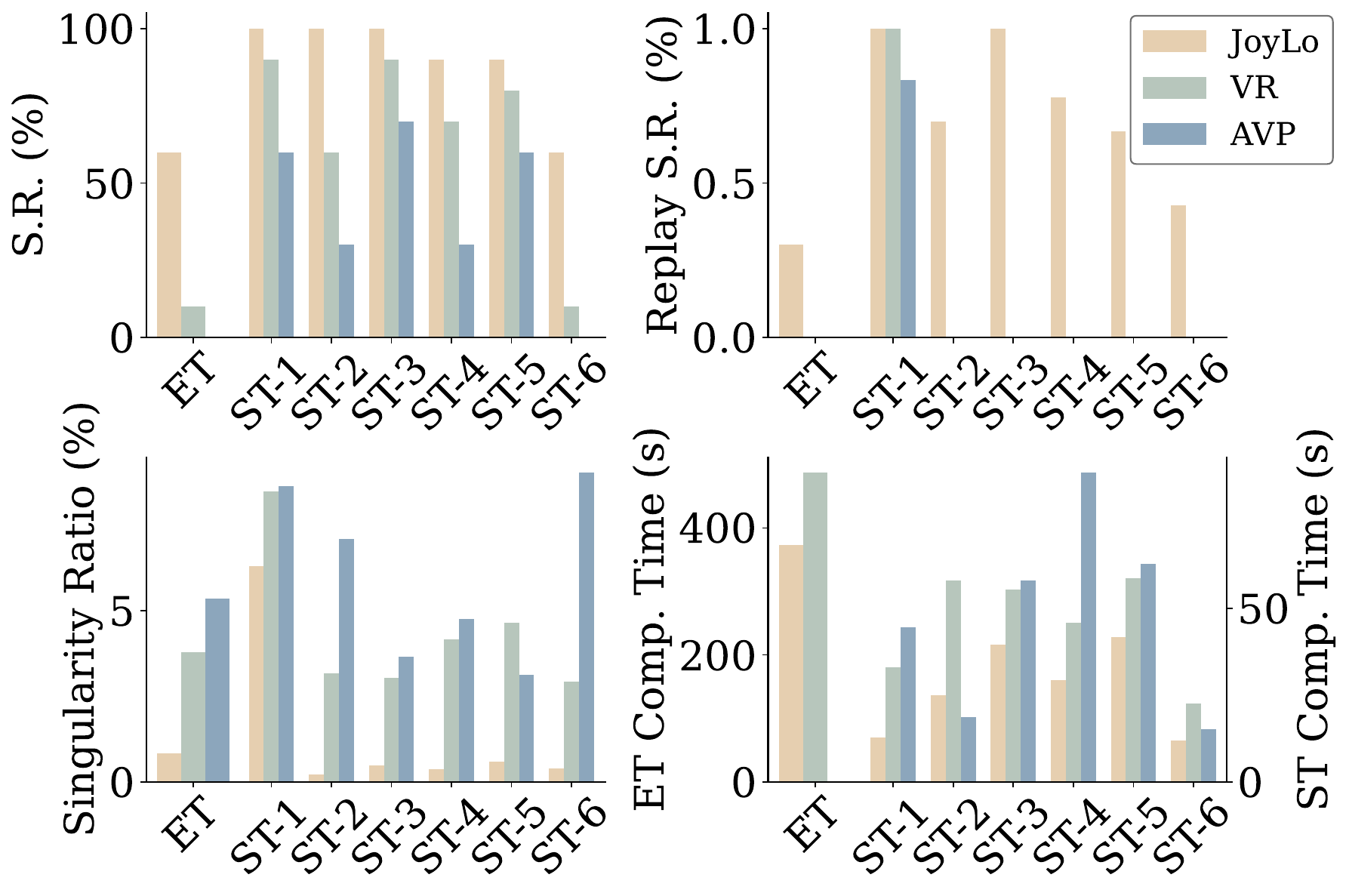}
    \vspace{-0.6cm}
    \caption{\textbf{User study results.} ``S.R.'' is success rate. ``ET Comp. Time'' and ``ST Comp. Time'' refer to entire and sub-task completion times.}
    \label{fig:user_study_result}
\end{wrapfigure}

As shown in Fig.~\ref{fig:user_study_result}, \interfacename achieves the highest success rate and fastest completion time across all interfaces.
It delivers a 5$\times$ higher task success rate and 23\% shorter median completion time than VR controllers, while no participants completed the entire task with Apple Vision Pro.
\interfacename particularly excels at articulated object manipulation (e.g., 67\% higher success in ``open dishwasher'' (ST-2) than VR controllers), enabling users to generate smooth and accurate actions, which is consistent with findings that leader-follower arm control improves fine-grained manipulation~\citep{zhao2023learning}.
It also significantly reduces sub-task times (e.g., 71\% faster navigation and 67\% faster bowl picking) compared to Apple Vision Pro, whose reliance on head movement for mobile base control leads to poor coordination and tracking~\citep{shaw2024bimanual}.
Moreover, \interfacename provides the highest data quality, achieving the lowest singularity ratio (78\% and 85\% lower than VR controllers and Apple Vision Pro, respectively) and consistently replaying successful trajectories.
Unlike IK-based methods that suffer from suboptimal IK solutions and jerky motions, JoyLo’s direct joint mapping and kinematic-twin arm constraints ensure smooth, stable whole-body teleoperation.
In user surveys (Fig.~\ref{appendix:fig:user_study_questionnaires}), all participants rated \interfacename the most user-friendly.
Although 70\% of participants initially believed IK-based interfaces would be more intuitive, after the study they unanimously preferred \interfacename.
This shift underscores a key distinction between tabletop data collection and mobile whole-body manipulation: while IK-based methods may suffice for static setups, they struggle to effectively control the mobile base and torso, making high-quality data collection much harder in mobile manipulation settings.

\begin{figure*}[t]
    \centering
    \includegraphics[width=\textwidth]{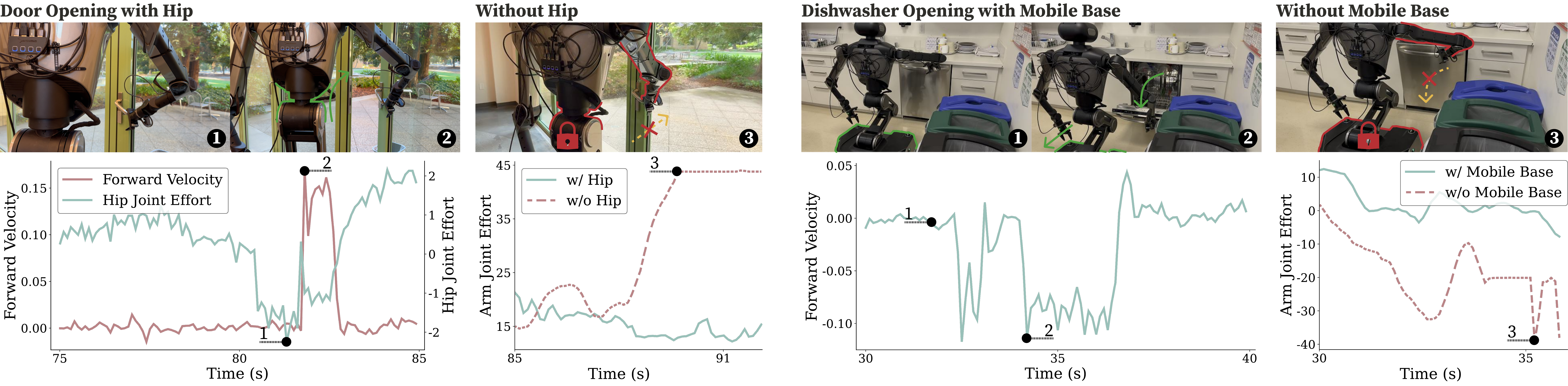}
    \vspace{-0.6cm}
    \caption{\textbf{Coordinated torso and mobile base movements enhance maneuverability.} \algoname policies use the hip and mobile base to open a door and dishwasher; if the torso or mobile base is locked, opening fails and arm joint effort surges, risking hardware damage.}
    \label{fig:emergent_behaviors}
    \vspace{-0.5cm}
\end{figure*}

\para{Coordinated torso and mobile base movements enhance maneuverability beyond stationary arms (\q{4}).}
As shown in Fig.~\ref{fig:emergent_behaviors}, coordinated whole-body movements are critical for tasks involving heavy articulated object interactions, such as ``open the door'' (ST-3) in ``take trash outside'' and ``open the dishwasher'' (ST-2) in ``clean house after a wild party.''
To open a door, the robot bends its hip forward while advancing the base to generate enough inertia; to open a dishwasher, it moves the base backward, using its whole body to pull the door open smoothly.
Without hip or base movement, both objects remain closed and the arm joint effort would surge, generating excessive force that is potentially harmful to the hardware.
Additional emergent behaviors such as failure recovery are showcased in videos on \webpage, demonstrating WB-VIMA’s robustness.

\section{Related Work}
\label{sec:related_work}
\textbf{\emph{Robots for Everyday Household Activities}}
Daily household activities have become a major focus for human-centered robotics~\citep{littman2022gathering,https://doi.org/10.1002/hbe2.117,10.1145/3328485,10.1145/3419764,doi:10.1126/science.aat8414}, with efforts mainly in: 1) defining benchmarks~\citep{kolve2017ai2thor,Puig_2018_CVPR,Shridhar_2020_CVPR,batra2020rearrangement,shridhar2020alfworld,srivastava2021behavior,pari2021surprising,Ehsani_2021_CVPR,Weihs_2021_CVPR,NEURIPS2021_021bbc7e,li2021igibson,li2022behaviork,heo2023furniturebench,yenamandra2023homerobot,shukla2024maniskillhab,nasiriany2024robocasa}, and 2) building robotic systems, usually with learning-based methods, to automate tasks~\citep{9196677,bahl2022humantorobot,bajracharya2023demonstrating,7139396,wang2023mimicplay,zhang2023noir,wu23tidybot,shi2023robocook,stone2023openworld,yang2023harmonic,fu2024mobile,xiong2024adaptive,jiang2024transic,yang2024equibot,dai2024automated,wu2024tidybot,shaw2024bimanual,black2024_0,hsu2024kinscene}.
Unlike field~\citep{Shamshiri2018Agricultural}, rescue~\citep{https://doi.org/10.1002/rob.21887}, or surgical robots~\citep{GOMES2011261}, household robots must generalize across diverse, complex home environments.
Prior works typically address either data collection or policy learning separately (Table~\ref{table:comparison}).
In contrast, \acronym offers a synergistic framework combining a low-cost, whole-body interface for data collection and a general, competent algorithm for whole-body visuomotor policy learning.
Moreover, many household tasks require \bimanual{bimanual} coordination and extensive end-effector \reachability{reachability}. Prior systems often rely on a single arm and lifting bodies~\citep{pari2021surprising,shafiullah2023bringing,yang2023harmonic}, whereas \acronym unleashes the mobile manipulation capabilities to perform broader real-world household tasks.

\textbf{\emph{Low-Cost Hardware for Robot Learning}}
Cost-effective hardware has accelerated robot learning, including: 1) low-cost robots---arms~\citep{zhao2023learning}, hands~\citep{7859295,shaw2023leap,shaw2024leap}, mobile manipulators~\citep{9196677,bajracharya2023demonstrating,fu2024mobile,xiong2024adaptive,wu2024tidybot}, and humanoids~\citep{5152516,7041473,9555790,https://doi.org/10.1002/rob.21702,10375199,liao2024berkeley,shi2025toddlerbot}; 2) teleoperation interfaces---puppeteering devices~\citep{wu2023gello,zhao2023learning,si2024tilde,shaw2024bimanual}, exoskeletons~\citep{9158331,fang2023airexo,yang2024ace}, and AR/VR devices~\citep{dass2024telemoma,iyer2024open,cheng2024opentelevision}; and 3) wearable or portable data collection devices~\citep{9126187,pmlr-v155-young21a,10341661,chi2024universal,wang2024dexcap,chen2024arcap,seo2024legato}.
Our \interfacename falls under teleoperation interfaces, providing a cost-effective, whole-body solution for mobile, dual-arm robots with torsos.
Unlike prior interfaces for stationary arms~\citep{wu2023gello,fang2023airexo} or mobile bases without independent torso control~\citep{fu2024mobile,shaw2024bimanual}, \interfacename enables efficient, untethered teleoperation of dual-arm mobile manipulators without needing a second operator.
Additionally, compared to common puppeteering devices~\citep{wu2023gello}, \interfacename offers rich haptic feedback via bilateral teleoperation without requiring force sensors~\citep{https://doi.org/10.1002/rob.21960,LI2016188} or extra real-robot arms~\citep{shridhar2024generative}.

\textbf{\emph{Learning Whole-Body Manipulation}}
Whole-body manipulation uses the full robot body, including arms~\citep{SMITH20121340,Vahrenkamp2011BimanualGP,doi:10.1126/science.aat8414,grannen2023stabilize,fu2024mobile}, torso~\citep{1285572,6630792,6202432,xu2025robopanoptes}, and base~\citep{9561315,yang2023harmonic,uppal2024spin,xiong2024adaptive,shah2024bumble,10328058,pmlr-v205-fu23a,liu2024visual,ha2024umi}, to interact with objects.
Traditional approaches rely on motion planning~\citep{371337,doi:10.1177/0278364913484072,doi:10.1177/02783640022067139,1642100,6202432,6630792,7041375,hsu2024kinscene}, while recent learning-based methods use reinforcement learning~\citep{9561315,honerkamp2022n2m2,pmlr-v205-fu23a,yang2023harmonic,hu2023causal,10328058,fu2024mobile,uppal2024spin,pan2024roboduet,arm2024pedipulate,liu2024visual,ha2024umi,he2024visual,zhang2024wococo}, behavior cloning~\citep{brohan2022rt1,fu2024mobile,fu2024humanplus,cheng2024opentelevision,wu2024tidybot,yang2024equibot,li2024okami,ze2024humanoid_manipulation,he2024omniho}, or large pretrained models~\citep{DBLP:conf/corl/IchterBCFHHHIIJ22,wu23tidybot,xu2023creative,stone2023openworld,jiang2024harmon,shah2024bumble,wu2024helpful}.
Our \algoname introduces a novel algorithm for learning whole-body manipulation on a high-DoF, wheeled, dual-arm robot with a torso.
Unlike prior methods that ignore action hierarchy~\citep{fu2024mobile,fu2024humanplus,wu2024tidybot} or embodiment interdependencies~\citep{pmlr-v205-fu23a,hu2023causal,pan2024roboduet}, \algoname explicitly models them through autoregressive whole-body action decoding, enabling coordinated policies for challenging real-world tasks.
Additionally, \algoname dynamically fuses multi-modal observations via visuomotor attention, extracting salient task-relevant information, which prior works~\citep{liu2024visual,yang2024equibot,ze2024humanoid_manipulation} often neglect.

\section{Conclusion}
\label{sec:conslusion}

This paper presents \acronym, a holistic framework for learning whole-body manipulation to tackle diverse real-world household tasks.
We identify three core capabilities essential for household activities: \bimanual{bimanual} coordination, stable \mobility{navigation}, and extensive end-effector \reachability{reachability}.
Achieving these with learning-based methods requires overcoming challenges in both data and modeling.
\acronym addresses them through two innovations: 1) \interfacename, a cost-effective whole-body interface for efficient data collection, and 2) \algoname, a novel algorithm that leverages embodiment hierarchy and models interdependent whole-body actions.
The \acronym system demonstrates strong performance across real-world household tasks with unmodified objects in natural, unstructured environments, marking a step toward greater autonomy and reliability in household robotics.

\section{Limitations}
\label{sec:limitations}

\begin{figure}[t]
\centering
\includegraphics[width=\textwidth]{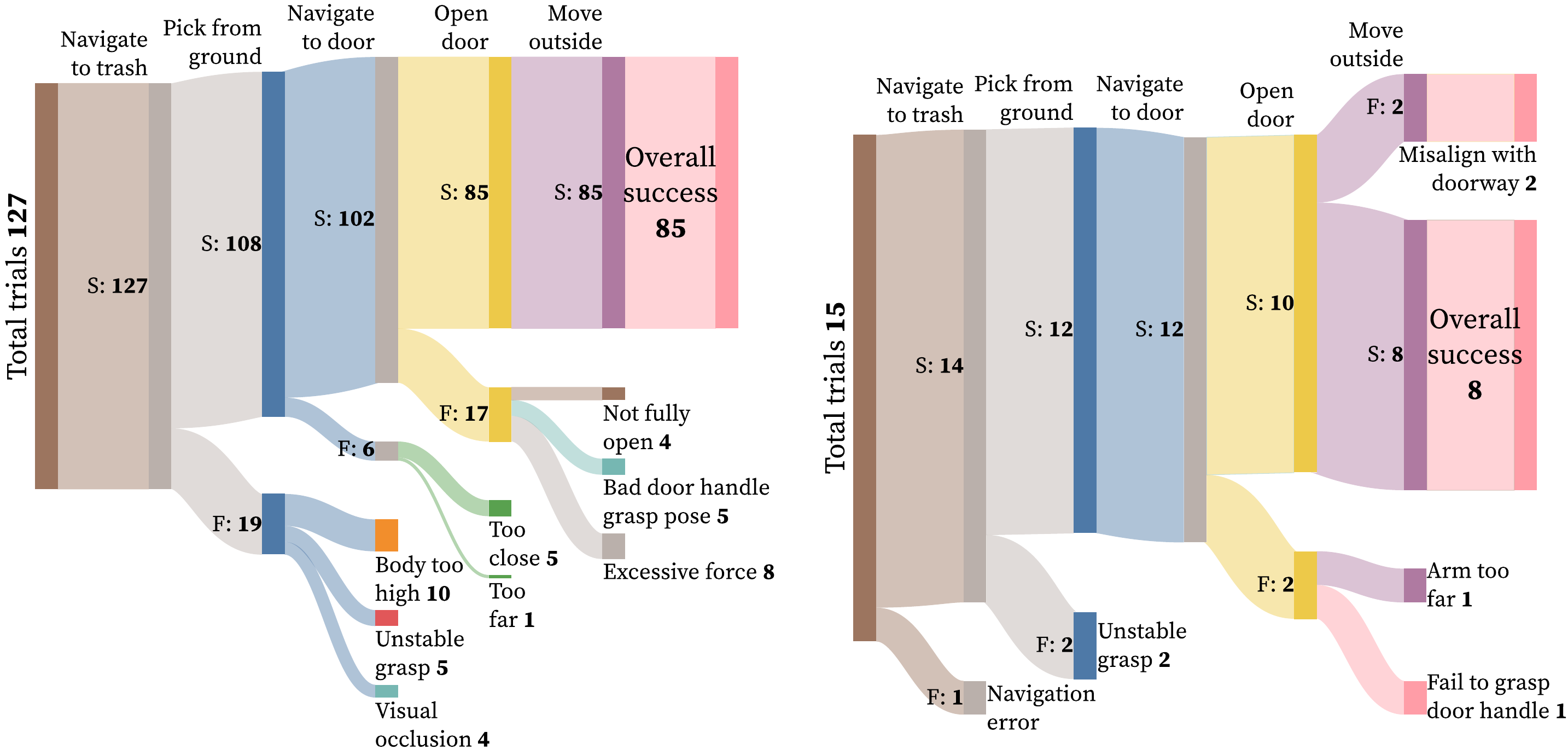}
\vspace{-0.5cm}
\caption{\textbf{Failure modes in the ``take trash outside'' task.} \textbf{Left:} Failure analysis during data collection using \interfacename. \textbf{Right:} Failure analysis during autonomous \algoname policy rollouts. ``S'' indicates number of successful trials. ``F'' indicates number of failed trials.}
\label{fig:failure_modes}
\end{figure}

While \acronym demonstrates strong performance across real-world household tasks, several limitations remain. In this section, we discuss limiting assumptions, analyze failure modes (Fig.~\ref{fig:failure_modes}), and suggest directions for future work.

\para{Mismatched camera field of view between robot and operator.}
During data collection with \interfacename, the operator observes the robot from a third-person perspective using their own vision. To collect data efficiently, they must position themselves to maintain a clear view of the workspace without appearing in the robot’s field of view. Additionally, the operator must ensure that target objects are visible to the robot’s cameras; otherwise, the resulting data will be partially observable, complicating policy training. Future work could incorporate active perception~\citep{cheng2024opentelevision,5968,xiong2025vision} so that the operator sees exactly what the robot sees.

\para{Compounding errors in long-horizon, multi-stage tasks.}
In complex tasks like ``clean house after a wild party,'' \algoname experiences compounding errors across multiple sub-tasks and over long horizons. While sub-task success rates remain high, these accumulated errors can significantly reduce overall task success. This limitation could be mitigated by learning on human correction data~\citep{ross2010reduction,jiang2024transic,wu2025robocopilot0} or integrating model-based task planning~\citep{liu2024learning} to improve robustness over extended execution.

\para{Imperfect point cloud observations.}
\algoname relies on point cloud data from onboard cameras, which can be degraded by lighting conditions or reflective surfaces. For example, policies trained on data collected during the day may not generalize well to nighttime environments due to visual discrepancies. Since our robot is equipped with stereo cameras, future work could incorporate FoundationStereo~\citep{wen2025foundationstereo0} to improve point cloud quality.

\para{Robot-specific training data.}
\algoname is trained on data collected exclusively with the R1 robot.
It is intriguing to explore how multi-embodiment data and cross-embodiment transfer can benefit the training~\citep{team2024octo,black2024_0,10611477,yang2024pushing,doshi2024scaling}.
The current dataset may also be insufficient for scene-level generalization. Future work could integrate large pre-trained models, such as VLA~\citep{brohan2023rt2,kim2024openvla,xu2024mobility}, to enhance scene understanding. Finally, it would be valuable to study how whole-body manipulation can benefit from synthetic data~\citep{mandlekar2023mimicgen,garrett2024skillmimicgen,li2025momagen} or human data~\citep{kareer2024egomimic,papagiannis2024rx,Grauman_2024_CVPR,nvidia2025gr00t}.

\acknowledgments{We thank Chengshu (Eric) Li, Wenlong Huang, Mengdi Xu, Ajay Mandlekar, Haoyu Xiong, Haochen Shi, Jingyun Yang, Toru Lin, Jim Fan, and the SVL PAIR group for their invaluable technical discussions.
We also thank Tianwei Li and the development team at Galaxea.ai for timely hardware support,
Yingke Wang for helping with the figures,
Helen Roman for processing hardware purchase,
Frank Yang, Yihe Tang, Yushan Sun, Chengshu (Eric) Li, Zhenyu Zhang, Haoyu Xiong for participating in user studies, and the Stanford Gates Building community for their patience and support during real-robot experiments. 
This work is in part supported by the Stanford Institute for Human-Centered AI (HAI), the Schmidt Futures Senior Fellows grant, NSF CCRI \#2120095, ONR MURI N00014-21-1-2801, ONR MURI N00014-22-1-2740, and ONR MURI N00014-24-1-2748.}

\clearpage
\bibliography{references}  %

\clearpage
\appendix
\renewcommand{\thefigure}{A.\arabic{figure}}
\renewcommand{\theequation}{A.\arabic{equation}}
\renewcommand{\thetable}{A.\Roman{table}}

\setcounter{figure}{0}
\setcounter{equation}{0}
\setcounter{table}{0}

\section{Robot Hardware Details}
\label{appendix:sec:hardware_details}
This section provides additional hardware details, including robot specifications, onboard sensors and computing, and the communication scheme.
\subsection{Robot Platform}
\label{appendix:sec:robot_platform}
We select the Galaxea R1 robot as our platform to meet the three critical capabilities essential for household tasks: \bimanual{bimanual} coordination, stable and precise \mobility{navigation}, and extensive end-effector \reachability{reachability}. As illustrated in Fig.\ref{fig:hardware_system}, the R1 robot features two 6-DoF arms mounted on a 4-DoF torso. Each arm is equipped with a parallel jaw gripper and has a maximum payload of \SI{5}{\kilogram}\footnote{All numbers related to the robot's hardware capabilities are based on our testing.}, making it well-suited for manipulating most objects encountered in daily household activities. The torso incorporates four revolute joints: two for waist rotation and hip bending, and two additional joints enabling knee-like motions. This design allows the robot to transition smoothly between standing and squatting positions, enhancing its reachability in household environments. By integrating the torso into the kinematic chain of the end-effectors, the R1 robot achieves an effective reach range from ground level to \SI{2}{\meter} vertically and up to \SI{2.06}{\meter} horizontally, covering the workspace shown in Fig.~\ref{fig:b1k_object_spatial_dist}. The arms and torso are controlled using joint impedance controllers, with target joint positions as inputs.

To ensure stable navigation in household environments, the robot’s torso is mounted on an omnidirectional mobile base, capable of moving in any direction on the ground plane at a maximum speed of \SI{1.5}{\meter\per\second}. Additionally, the base can independently execute yaw rotations at a maximum angular speed of \SI{3}{\radian\per\second}. This mobility is powered by three wheel motors and three steering motors. With a \SI{30}{\milli\meter} ground clearance, the mobile base can traverse most household terrains. It also achieves horizontal accelerations of up to \SI{2.5}{\meter\per\square\second}, enhancing maneuverability for tasks that require simultaneous movement and manipulation, such as opening doors (Fig.~\ref{fig:emergent_behaviors}). The mobile base is controlled via velocity commands corresponding to its three degrees of freedom on the ground plane: forward motion, lateral motion, and yaw rotation.

For perception, we equip the R1 robot with a suite of onboard sensors, including a stereo ZED 2 RGB-D camera as the head camera, two stereo ZED-Mini RGB-D cameras as wrist cameras, and a RealSense T265 tracking camera for visual odometry. All RGB-D cameras operate at \SI{60}{\hertz}, streaming rectified RGB and depth images. The cameras' poses are updated at \SI{500}{\hertz} via the robot’s forward kinematics, enabling the effective fusion of sensory data from all three cameras. This integration supports high-fidelity global and ego-centric 3D perception, such as colored point-cloud observations. Simultaneously, the visual odometry system operates at \SI{200}{\hertz}, providing real-time velocity and acceleration estimates of the mobile base, which is critical feedback for learning precise velocity control for the mobile base.

\subsection{Hardware Specifications}
\begin{figure}[t]
    \centering
    \begin{subfigure}{\textwidth}
    \includegraphics[width=\textwidth]{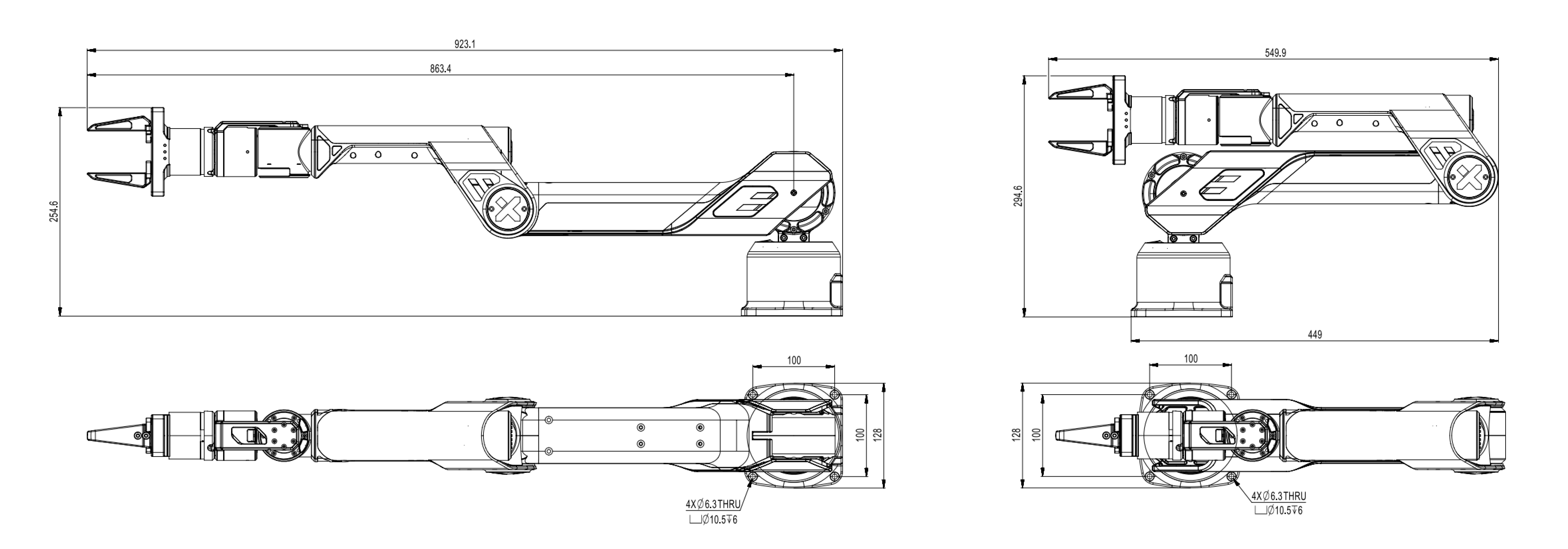}
    \caption{}
    \label{appendix:fig:arm_digram}
    \end{subfigure}

    \begin{subfigure}{0.3\textwidth}
    \includegraphics[width=\textwidth]{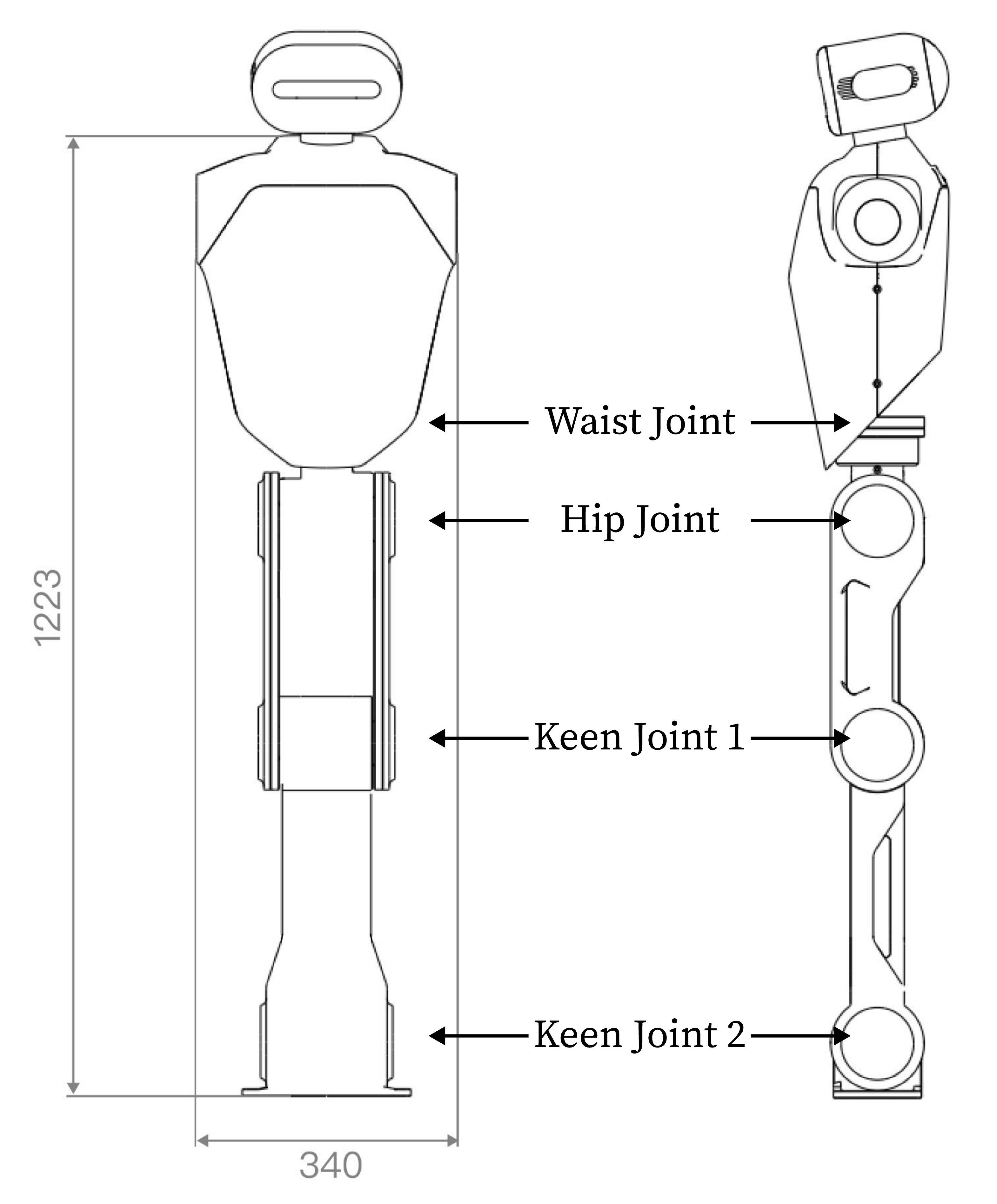}
    \caption{}
    \label{appendix:fig:torso_digram}
    \end{subfigure}\hfill{}\begin{subfigure}{0.6\textwidth}
            \includegraphics[width=\textwidth]{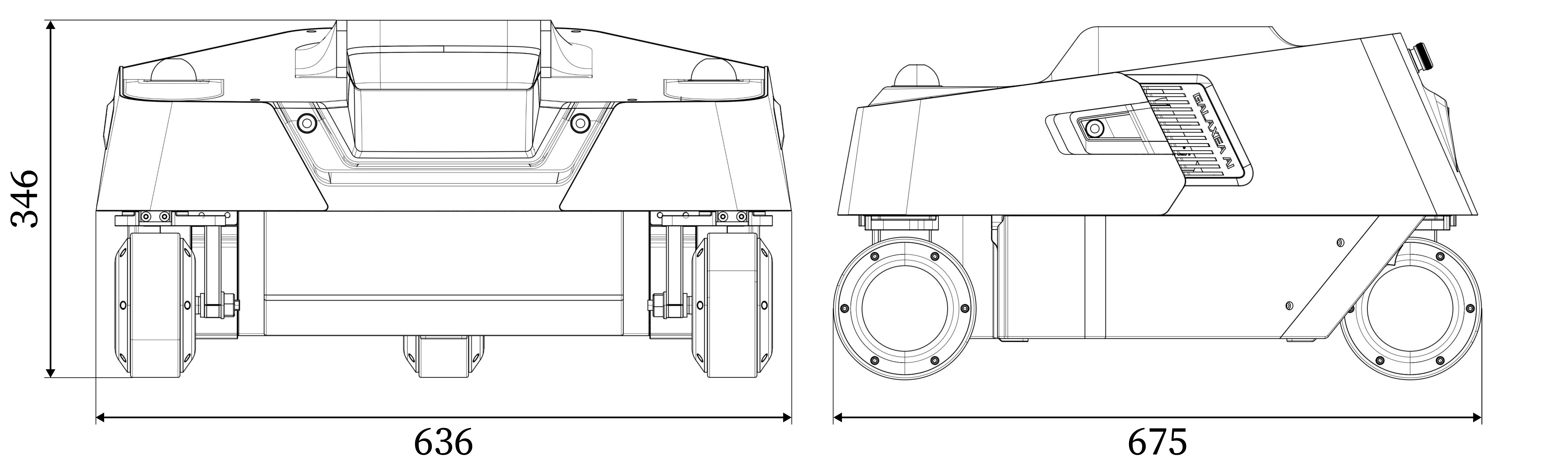}
    \caption{}
    \label{appendix:fig:base_digram}

    \end{subfigure}
    \caption{\textbf{Robot diagrams.} \textbf{(a):} Each arm has six DoFs and a parallel jaw gripper. \textbf{(b):} The torso features four revolute joints for waist rotation, hip bending, and knee-like motions. \textbf{(c):} The wheeled, omnidirectional mobile base is equipped with three steering motors and three wheel motors.}
\end{figure}

\subsubsection{Arms} The Galaxea R1 robot has two 6-DoF arms, each equipped with a parallel jaw gripper. As shown in Fig.~\ref{appendix:fig:arm_digram}, each arm has a \SI{128}{\milli\meter} width and a \SI{923}{\milli\meter} full reach.
The arms are mirrored on the robot and are controlled via a joint impedance controller, receiving target joint positions as inputs.
We set the following impedance gains: $\mathbf{K_p} = [140, 200, 120, 20, 20, 20]$ and $\mathbf{K_d} = [10, 50, 5, 1, 1, 0.4]$.
Each gripper has a stroke range from \SI{0}{\milli\meter} (fully closed) to \SI{100}{\milli\meter} (fully open), with a rated gripping force of \SI{100}{\newton}.
The grippers are controlled by specifying a target opening width, which is converted into the required motor current.

\subsubsection{Torso}
The torso consists of four revolute joints: two joints for waist rotation and hip bending, and two additional joints for knee-like motions.
As shown in Fig.~\ref{appendix:fig:torso_digram}, the torso has a \SI{340}{\milli\meter} width and a \SI{1223}{\milli\meter} height (excluding the head) when fully extended.
Table~\ref{appendix:table:torso_motor_spec} lists the motor specifications.
\begin{table}[ht]
\centering
\caption{\textbf{Torso motor specifications.}}
\label{appendix:table:torso_motor_spec}
\begin{tabular}{@{}cc@{}}
\toprule
\textbf{Parameter}      & \textbf{Value}            \\ \midrule
Waist Joint Range (Yaw) & $\pm$ \SI{3.05}{\radian} (\SI{175}{\degree})                       \\
Hip Joint Range (Pitch) & \SI{-2.09}{\radian} (\SI{-120}{\degree}) $\sim$  \SI{1.83}{\radian} (\SI{105}{\degree})                     \\
Knee Joint 1 Range      & \SI{-2.79}{\radian} (\SI{-160}{\degree}) $\sim$  \SI{2.53}{\radian} (\SI{145}{\degree})                       \\
Knee Joint 2 Range      & \SI{-1.13}{\radian} (\SI{-65}{\degree}) $\sim$  \SI{1.83}{\radian} (\SI{105}{\degree})                          \\
Rated Motor Torque            & \SI{108}{\newton\meter} \\ 
Maximum Motor Torque            & \SI{304}{\newton\meter} \\ 
\bottomrule
\end{tabular}
\end{table}

\subsubsection{Mobile Base}
As illustrated in Fig.~\ref{appendix:fig:base_digram}, the mobile base is wheeled and omnidirectional, equipped with three steering motors and three wheel motors.
The base can move in any direction on the ground plane and perform yaw rotations. It is controlled via a velocity controller with 3-DoF inputs corresponding to forward velocity (x-axis), lateral velocity (y-axis), and rotation velocity (z-axis).
Performance parameters are listed in Table~\ref{appendix:table:mobile_base_spec}.

\begin{table}[ht]
\centering
\caption{\textbf{Mobile base specifications.}}
\label{appendix:table:mobile_base_spec}
\begin{tabular}{@{}cc@{}}
\toprule
\textbf{Parameter}              & \textbf{Value} \\ \midrule
Forward Velocity Limit          &   $\pm$ \SI{1.5}{\meter\per\second}             \\
Lateral Velocity Limit          &     $\pm$ \SI{1.5}{\meter\per\second}             \\
Yaw Rotation Velocity Limit     &     $\pm$ \SI{3}{\radian\per\second}             \\
Forward Acceleration Limit      &  $\pm$ \SI{2.5}{\meter\per\square\second}              \\
Lateral Acceleration Limit      &   $\pm$ \SI{1.0}{\meter\per\square\second}             \\
Yaw Rotation Acceleration Limit &   $\pm$ \SI{1.0}{\radian\per\square\second}             \\ \bottomrule
\end{tabular}
\end{table}

\subsection{Onboard Sensors and Computing}
As shown in Fig.~\ref{fig:hardware_system}, the robot is equipped with several onboard sensors: a ZED 2 RGB-D camera (head camera), two ZED-Mini RGB-D cameras (wrist cameras), and a RealSense T265 tracking camera (visual odometry).
Camera configurations are provided in Table~\ref{appendix:table:camera_config}.

\begin{table}[ht]
\centering
\caption{\textbf{Configurations for the ZED RGB-D cameras and RealSense T265 tracking camera.}}
\label{appendix:table:camera_config}
\begin{tabular}{@{}cc@{}}
\toprule
\textbf{Parameter}     & \textbf{Value} \\ \midrule
\multicolumn{2}{c}{RGB-D Cameras}       \\ \midrule
Frequency              & \SI{60}{\hertz}             \\
Image Resolution       & 1344$\times$376       \\
ZED Depth Mode         & \texttt{PERFORMANCE}    \\
Head Camera Min Depth  & 0.2            \\
Head Camera Max Depth  & 3              \\
Wrist Camera Min Depth & 0.1            \\
Wrist Camera Max Depth & 1              \\ \midrule
\multicolumn{2}{c}{Tracking Camera}     \\ \midrule
Odometry Frequency     & \SI{200}{\hertz}            \\ \bottomrule
\end{tabular}
\end{table}

\begin{wrapfigure}[14]{r}{0.5\textwidth}
    \vspace{-0.5cm}
    \centering
    \includegraphics[width=0.5\textwidth]{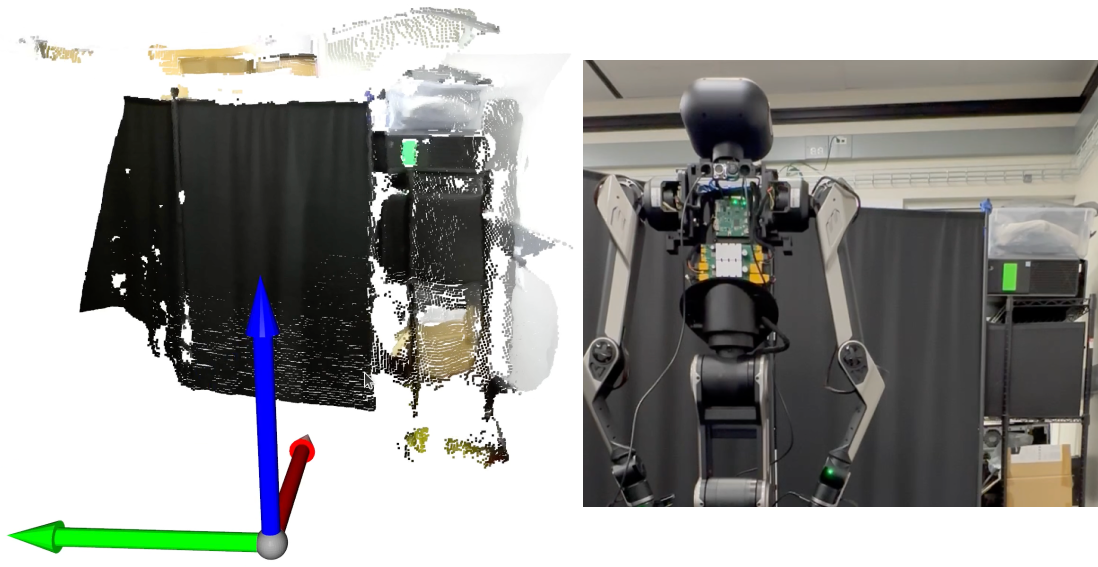}
    \caption{\textbf{Visualization of the fused, ego-centric colored point clouds.} \textbf{Left:} The colored point cloud observation, aligned with the robot’s coordinate frame. \textbf{Right:} The robot’s orientation and its surrounding environment.}
    \label{appendix:fig:fused_pcd_viz}
\end{wrapfigure}

The three RGB-D cameras stream colored point clouds at \SI{60}{\hertz}, obtained from rectified RGB images and aligned depth images. These point clouds are fused into a common robot base frame.
For each point cloud in the camera frame $\mathbf{P}^{camera}$, where $camera \in \text{all cameras} = \{\text{head}, \text{left wrist}, \text{right wrist}\}$, the transformation from the robot base frame to camera frames is computed using forward kinematics at \SI{500}{\hertz}.
Denote rotation matrices as $\mathbf{R}^{camera} \in \mathbb{R}^{3 \times 3}$ and translations as $\mathbf{t}^{camera} \in \mathbb{R}^{3 \times 1}$, the fused, ego-centric point cloud $\mathbf{P}^\text{ego-centric}$ is computed as $\mathbf{P}^\text{ego-centric} = \bigcup_{camera}^\text{all cameras}\mathbf{P}^{camera}\left(\mathbf{R}^{camera}\right)^\intercal + \left(\mathbf{t}^{camera}\right)^\intercal$.
An example of the fused ego-centric colored point cloud is shown in Fig.~\ref{appendix:fig:fused_pcd_viz}. The point cloud is then spatially cropped and downsampled using farthest point sampling (FPS)~\citep{GONZALEZ1985293, qi2017pointnet, han2023quickfps}.

The RealSense T265 tracking camera provides 6D velocity and acceleration feedback at \SI{200}{\hertz}.
It is mounted on the back of the mobile base using a custom-designed camera mount.

The R1 robot is equipped with an NVIDIA Jetson Orin, dedicated to running cameras and processing observations at a high rate.

\subsection{Communication Scheme}
The robot communicates with a workstation via the Robot Operating System (ROS).
Each camera operates as an individual ROS node. The workstation runs the master ROS node, which subscribes to robot state nodes and camera nodes, and issues control commands via ROS topics.
To reduce latency, a local area network (LAN) is established between the workstation and the robot.

\section{\interfacename Details}
This section provides details on \interfacename, including its hardware components, controller implementation, and data collection process.

\subsection{Hardware Components}
\begin{wrapfigure}[8]{r}{0.4\textwidth}
    \vspace{-0.4cm}
    \centering
    \includegraphics[width=0.4\textwidth]{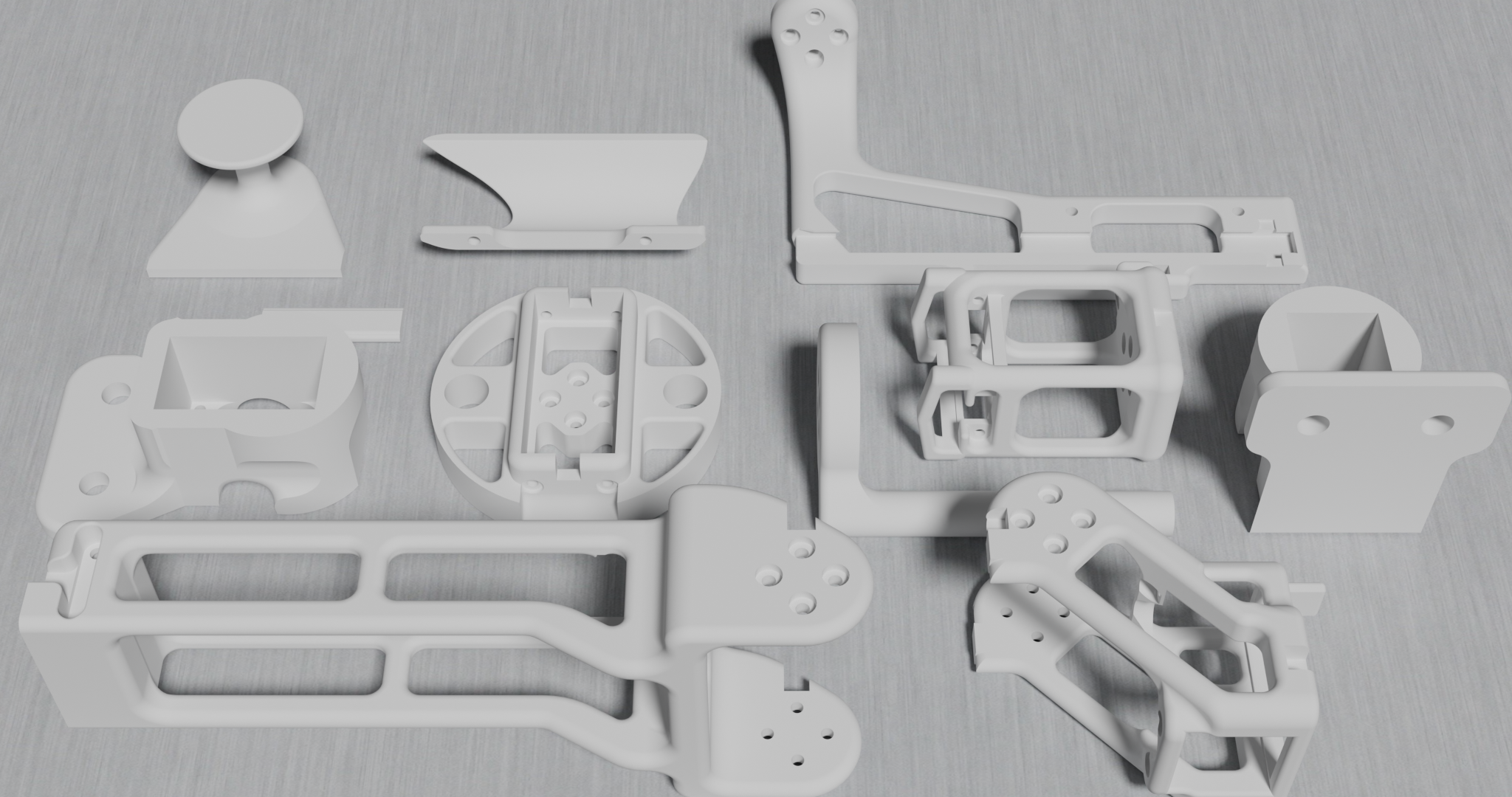}
    \caption{\textbf{Individual \interfacename links.}}
    \label{appendix:fig:joylo_disassembled}
\end{wrapfigure}

The \interfacename system consists of 3D-printable arm links, low-cost Dynamixel motors, and off-the-shelf Joy-Con controllers.
The individual arm links are shown in Fig.~\ref{appendix:fig:joylo_disassembled}. Using a Bambu Lab P1S 3D printer, we printed two arms in \SI{13}{\hour}, consuming \SI{317}{\gram} of PLA filament.
The bill of materials is listed in Table~\ref{appendix:table:bom}.
Once assembled, we use the official Dynamixel SDK to read motor states at \SI{400}{\hertz} - \SI{500}{\hertz}.
The Joy-Cons connect to the workstation via Bluetooth, communicating at \SI{66}{\hertz}.

\begin{table*}[ht]
\centering
\caption{\textbf{\interfacename bill of materials.}}
\label{appendix:table:bom}
\resizebox{1\textwidth}{!}{
\begin{tabular}{@{}cllcccc@{}}
\toprule
\textbf{Item No.} & \multicolumn{1}{c}{\textbf{Part Name}} & \multicolumn{1}{c}{\textbf{Description}}                     & \textbf{Quantity} & \textbf{Unit Price (\$)} & \textbf{Total Price (\$)} & \textbf{Supplier} \\ \midrule
1                 & Dynamixel XL330-M288-T                 & \interfacename arm joint motors                                       & 16                & 23.90                & 382.40               & Dynamixel         \\
2                 & Nintendo Joy-Con                       & \interfacename hand-held controllers                                  & 1                 & 70                   & 70                   & Nintendo          \\
3                 & Dynamixel U2D2                         & USB communication converter for controlling Dynamixel motors & 1                 & 32.10                & 32.10                & Dynamixel         \\
4                 & 5V DC Power Supply                     & Power supply for Dynamixel motors                            & 1                 & \textless{}10        & \textless{}10        & Various           \\
5                 & 3D Printer PLA Filament                & PLA filament for 3D printing \interfacename arm links                 & 1                 & $\sim$5              & $\sim$5              & Various           \\ \midrule
\multicolumn{7}{l}{\textbf{Total Cost: $\sim$\$499.5}}                                                                                                                                                           
\end{tabular}
}
\end{table*}

\subsection{Controller Implementation}
We provide an intuitive, real-time Python-based controller to operate \interfacename with the R1 robot.
As illustrated in Pseudocode~\ref{appendix:controller_code_snippet}, the controller includes a joint impedance controller for the torso and arms with target joint positions as inputs, and a velocity controller for the mobile base with target base velocities as inputs.
Control commands are converted into waypoints and sent to the robot via ROS topics at  \SI{100}{\hertz}, which we find to be sufficient in practice.

To enable bilateral teleoperation of \interfacename arms as discussed in Sec.~\ref{sec:joylo}, we implement a joint impedance controller using current-based control, where force is proportional to motor current.
We set proportional gains $\mathbf{K_p} = [0.5, 0.5, 0.5, 0.5, 0.5, 0.5]$ and derivative gains $\mathbf{K_d} = [0.01, 0.01, 0.01, 0.01, 0.01, 0.01]$.
To ensure sufficient stall torque for load-bearing joints in the \interfacename arms, such as the shoulder joints, the two low-cost Dynamixel motors are coupled together, as illustrated in Fig.~\ref{fig:hardware_system}.

\subsection{Data Collection}
During data collection, the robot operates at \SI{100}{\hertz}, while samples are recorded at \SI{10}{\hertz}.
Functional buttons on the right Joy-Con (Fig.~\ref{fig:hardware_system}) control start, pause, save, and discard actions.
Recorded data includes RGB images, depth images, point clouds, joint states, odometry, and action commands.
\begin{adjustbox}{minipage=\textwidth,center}
\centering
\begin{lstlisting}[language=Python,caption={\textbf{Python interface for the R1 robot controller.}},label={appendix:controller_code_snippet}]
from brs_ctrl.robot_interface import R1Interface

# instantiate the controller
robot = R1Interface(...)
# send a control command
robot.control(
    # the torso and arms commands are target joint positions
    arm_cmd={
        "left": left_arm_target_q,
        "right": right_arm_target_q,
    },
    gripper_cmd={
        "left": left_gripper_target_width,
        "right": left_gripper_target_width,
    },
    torso_cmd=torso_target_q,
    # the mobile base commands are target velocities
    base_cmd=mobile_base_target_velocity,
)
\end{lstlisting}
\end{adjustbox}

\section{Model Architectures, Policy Training, and Deployment Details}
\label{appendix:sec:policy_training_details}
This section provides details on \algoname and baseline model architectures, policy training, and real-robot deployment.

\subsection{Preliminaries}
\para{Problem Formulation}
We formulate robot manipulation as a Markov Decision Process (MDP) $\mathcal{M} \coloneqq \left( \mathcal{S}, \mathcal{A}, \mathcal{T}, \rho_0, R \right)$, where $s \in \mathcal{S}$ represents states, $a \in \mathcal{A}$ represents actions, $\mathcal{T}$ is the transition function, $\rho_0$ is the initial state distribution, and $R$ is the reward function~\citep{Sutton1998}.
A policy $\pi_\theta$, parameterized by $\theta$, learns the mapping $\mathcal{S} \rightarrow \mathcal{A}$.

\para{Denoising Diffusion for Policy Learning}
A denoising diffusion probabilistic model (DDPM)~\citep{NEURIPS2020_4c5bcfec,pmlr-v139-nichol21a,pmlr-v37-sohl-dickstein15} represents the data distribution $p(x^0)$ as the reverse denoising process of a forward noising process $q(x^k \vert x^{k - 1})$, where Gaussian noise is iteratively applied.
Given a noisy sample $x^k$ and timestep $k$ in the forward process, a neural network $\epsilon_\theta (x^k, k)$, parameterized by $\theta$, learns to predict the applied noise $\epsilon$.
Starting with a random sample $x^K \sim \mathcal{N}(0, I)$, the reverse denoising process is described as
\begin{equation}
    x^{k - 1} \sim \mathcal{N}\left(\mu_k\left(x^k, \epsilon_\theta \left(x^k, k\right) \right), \sigma_k^2I\right),
\end{equation}
where $\mu_k(\cdot)$ maps the noisy sample $x^k$ and the predicted noise $\epsilon_\theta$ to the mean of the next distribution, and $\sigma^2_k$ is the variance obtained from a predefined schedule for $k = 1, \ldots, K$.
Recently, DDPMs have been utilized to model policies $\pi_\theta$, where the denoising network $\epsilon_\theta ( a^k \vert s, k)$ is trained through behavior cloning~\citep{wang2022diffusion,chi2023diffusion,wang2024onestep}.

\subsection{\algoname Architecture}
\subsubsection{Observation Encoder}
As introduced in Sec.~\ref{sec:method:wb_vima}, there are two types of observation tokens: the point-cloud token $\mathbf{E}^{\text{pcd}}$ and the proprioceptive token $\mathbf{E}^\text{prop}$.
A colored point-cloud observation is denoted as $\mathbf{P}^\text{colored pcd} \in \mathbb{R}^{N_{\text{pcd}} \times 6}$, where $N_{\text{pcd}}$ is the number of points in the point cloud. Each point contains six channels: three for RGB values and three for spatial coordinates. 
To encode point-cloud tokens, RGB values are normalized to $[0,1]$ by dividing by 255; spatial coordinates are normalized to $[-1,1]$ by dividing by task-specific spatial limits; finally, a PointNet encoder~\citep{qi2016pointnet} processes the point cloud.
Proprioceptive observations include the mobile base velocity $v_\text{mobile base} \in \mathbb{R}^3$, torso joint positions $q_\text{torso} \in \mathbb{R}^{4}$, arms joint positions $q_\text{arms} \in \mathbb{R}^{12}$, and gripper widths $q_\text{grippers} \in \mathbb{R}^{2}$.
These values are concatenated and processed through an MLP.
Model hyperparameters for the PointNet and proprioception MLP are listed in Table~~\ref{appendix:table:pointnet_prop_mlp}.

\begin{table}[ht]
\centering
\caption{\textbf{Hyperparameters for PointNet and the proprioception MLP.}}
\label{appendix:table:pointnet_prop_mlp}
\begin{tabular}{@{}cclc@{}}
\toprule
\textbf{Hyperparameter} & \textbf{Value}            & \multicolumn{1}{c}{\textbf{Hyperparameter}} & \textbf{Value} \\ \midrule
\multicolumn{2}{c|}{PointNet}                       & \multicolumn{2}{c}{Prop. MLP}                                \\ \midrule
$N_\text{pcd}$                  & \multicolumn{1}{c|}{4096} & Input Dim                                   & 21             \\
Hidden Dim              & \multicolumn{1}{c|}{256}  & Hidden Dim                                  & 256            \\
Hidden Depth            & \multicolumn{1}{c|}{2}    & Hidden Depth                                & 3              \\
Output Dim              & \multicolumn{1}{c|}{256}  & Output Dim                                  & 256            \\
Activation              & \multicolumn{1}{c|}{GELU} & Activation                                  & ReLU           \\ \bottomrule
\end{tabular}
\end{table}

\subsubsection{Multi-Modal Observation Attention}
To effectively fuse multi-modal observations, \algoname employs a multi-modal observation attention network---a transformer decoder that applies causal self-attention over the input sequence:
$\mathbf{S} = [\mathbf{E}^\text{pcd}_{t - T_o + 1}, \mathbf{E}^\text{prop}_{t - T_o + 1}, \mathbf{E}^\text{a}_{t - T_o + 1},\ldots, \mathbf{E}^\text{pcd}_{t}, \mathbf{E}^\text{prop}_{t}, \mathbf{E}^\text{a}_{t}  ] \in \mathbb{R}^{3T_o \times E}$,
where $T_o$ is the observation window size, $E$ is the token dimension, and $\mathbf{E}^\text{a}$ represents the action readout token.
The transformer decoder's hyperparameters are listed in Table~\ref{appendix:table:xf_decoder_hyperparameters}. Action readout tokens are passive and do not influence the transformer output; they only attend to previous observation tokens to maintain causality.  
The final action readout token at time step $t$, $\mathbf{E}^a_t$, is used for autoregressive whole-body action decoding. We use an observation window size of $T_o = 2$ for all methods.

\begin{table}[ht]
\centering
\caption{\textbf{Hyperparameters for the transformer decoder used in multi-modal observation attention.}}
\label{appendix:table:xf_decoder_hyperparameters}
\begin{tabular}{@{}cc@{}}
\toprule
\textbf{Hyperparameter} & \textbf{Value} \\ \midrule
Embed Size              & 256            \\
Num Layers              & 2              \\
Num Heads               & 8              \\
Dropout Rate            & 0.1            \\
Activation              & GEGLU~\citep{shazeer2020glu}          \\ \bottomrule
\end{tabular}
\end{table}

\subsubsection{Autoregressive Whole-Body Action Decoding}
As discussed in Sec.~\ref{sec:method:wb_vima}, \algoname jointly learns three independent denoising networks for the mobile base, torso, and arms, denoted as $\epsilon_{\text{base}}$, $\epsilon_{\text{torso}}$, and $\epsilon_{\text{arms}}$, respectively.
Each denoising network is implemented using a UNet~\citep{ronneberger2015unet}, with hyperparameters listed in Table~\ref{appendix:table:unet}.
The denoising process follows three sequential steps.
First, the mobile base denoising network $\epsilon_{\text{base}}$ takes the action readout token $\mathbf{E}^a$ as input and predicts future mobile base actions $\mathbf{a}_\text{base} \in \mathbb{R}^{T_a \times 3}$.
Subsequently, the torso denoising network $\epsilon_{\text{torso}}$ takes $\mathbf{E}^a$ and $\mathbf{a}_\text{base}$ as input and predicts future torso actions $\mathbf{a}_\text{torso} \in \mathbb{R}^{T_a \times 4}$.
Finally, the arms denoising network $\epsilon_{\text{arms}}$ takes $\mathbf{E}^a$, $\mathbf{a}_\text{base}$, and $\mathbf{a}_\text{torso}$ as input and predicts future arm and gripper actions $\mathbf{a}_\text{arms} \in \mathbb{R}^{T_a \times 14}$.
Here $T_a$ is the action prediction horizon, and we use $T_a = 8$ hereafter.
To ensure low-latency inference, denoising starts from the encoded action readout tokens, meaning the observation encoders and transformer run only once per inference call.

\begin{table}[ht]
\centering
\caption{\textbf{Hyperparameters for the UNet models used for denoising.}}
\label{appendix:table:unet}
\begin{tabular}{@{}cc@{}}
\toprule
\textbf{Hyperparameter} & \textbf{Value} \\ \midrule
Hidden Dim              & [64,128]         \\
Kernel Size             & 2              \\
GroupNorm Num Groups    & 5              \\
Diffusion Step Embd Dim & 8              \\ \bottomrule
\end{tabular}
\end{table}

\subsection{Baselines Architectures}
We provide details on baseline methods DP3~\citep{ze20243d}, RGB-DP~\citep{chi2023diffusion}, and ACT~\citep{zhao2023learning}.
DP3 uses the same PointNet encoder as \algoname (Table~\ref{appendix:table:pointnet_prop_mlp}), but ignores RGB channels. Proprioceptive features are processed through the same MLP encoder. Encoded features are concatenated and passed through a fusion MLP with two hidden layers and 512 hidden units. A UNet denoising network (Table~\ref{appendix:table:unet}) predicts a flattened 21-DoF whole-body action trajectory.
RGB-DP is similar to DP3 but uses a pre-trained ResNet-18~\citep{he2015deep} as the vision encoder. The last classification layer is replaced with a 512-dimensional output layer for policy learning. We use the recommended hyperparameters provided in \citet{zhao2023learning} for ACT.

\subsection{Policy Training Details}
Policies are trained using the AdamW optimizer~\citep{loshchilov2017decoupled}, with hyperparameters in Table~\ref{appendix:table:train_hyperparameters}.  
90\% of collected data is used for training, and 10\% is reserved for validation.
Policies are trained for equal steps, using the last checkpoint for evaluation.
During training, we use the DDPM noise scheduler~\citep{NEURIPS2020_4c5bcfec,pmlr-v139-nichol21a,pmlr-v37-sohl-dickstein15} with 100 denoising steps.
During evaluation and inference, we use the DDIM noise scheduler~\citep{song2020denoising} with 16 denoising steps.
Training is performed using Distributed Data Parallel (DDP) on NVIDIA GPUs, including RTX A5000, RTX 4090, and A40.

\begin{table}[ht]
\centering
\caption{\textbf{Training hyperparameters.}}
\label{appendix:table:train_hyperparameters}
\begin{tabular}{@{}cc@{}}
\toprule
\textbf{Hyperparameter}                              & \textbf{Value}              \\ \midrule
Learning Rate                                        & $7 \times 10^{-4}$                        \\
Weight Decay                                         & 0.1                         \\
Learning Rate Warm Up Steps                          & 1000                        \\
\multicolumn{1}{l}{Learning Rate Cosine Decay Steps} & \multicolumn{1}{l}{300,000} \\
Minimal Learning Rate                                & $5 \times 10^{-6}$                        \\ \bottomrule
\end{tabular}
\end{table}

\subsection{Policies Deployment Details}
During deployment, observations from the robot's onboard sensors are transmitted to a workstation, where policy inference is performed, and the resulting actions are sent back for execution.
To minimize latency, we implement asynchronous policy inference.
Concretely, policy inference runs continuously in the background.
When switching to a new predicted trajectory, the initial few actions are discarded to compensate for inference latency.  
This ensures non-blocking execution, preventing delays caused by observation acquisition and controller execution.

\section{Task Definition and Evaluation Details}
\label{appendix:sec:experiment_settings}
This section provides detailed task definitions, generalization conditions, and evaluation protocols.

\subsection{Task Definition}
\label{appendix:sec:task_definition}
\textit{Activity 1 \textbf{Clean House After a Wild Party} (Fig.~\ref{fig:pull} First Row)}:
Starting in the living room, the robot navigates to a dishwasher in the kitchen (\textbf{ST-1}) and opens it (\textbf{ST-2}). It then moves to a gaming table (\textbf{ST-3}) to collect bowls (\textbf{ST-4}). Finally, the robot returns to the dishwasher (\textbf{ST-5}), places the bowls inside, and closes it (\textbf{ST-6}).
Stable and accurate \mobility{navigation} is the most critical capability for this task.
We collect 138 demonstrations, with an average human completion time of \SI{210}{\second}.
We randomize the starting position of the robot, bowl instances and their placements, and distractors on the table.

\textit{Activity 2 \textbf{Clean the Toilet} (Fig.~\ref{fig:pull} Second Row)}:
In a restroom, the robot picks up a sponge placed on a closed toilet (\textbf{ST-1}), opens the toilet cover (\textbf{ST-2}), cleans the seat (\textbf{ST-3}), closes the cover (\textbf{ST-4}), and wipes it (\textbf{ST-5}). The robot then moves to press the flush button (\textbf{ST-6}).
Extensive end-effector \reachability{reachability} is the most critical capability for this task.
We collect 103 demonstrations, with an average human completion time of \SI{120}{\second}.
We randomize the robot starting position, sponge instances, and placements.

\textit{Activity 3 \textbf{Take Trash Outside} (Fig.~\ref{fig:pull} Third Row)}:
The robot navigates to a trash bag in the living room, picks it up (\textbf{ST-1}), carries it to a closed door (\textbf{ST-2}), opens the door (\textbf{ST-3}), moves outside, and deposits the trash bag into a trash bin (\textbf{ST-4}).
Stable and accurate \mobility{navigation} is the most critical capability for this task.
We collect 122 demonstrations, with an average human completion time of \SI{130}{\second}.
We randomize the robot starting position and the placement of the trash bag.

\textit{Activity 4 \textbf{Put Items onto Shelves} (Fig.~\ref{fig:pull} Fourth Row)}:
In a storage room, the robot lifts a box from the ground (\textbf{ST-1}), moves to a four-level shelf, and places the box on the appropriate level based on available space (\textbf{ST-2}).
Extensive end-effector \reachability{reachability} is the most critical capability for this task.
We collect 100 demonstrations, with an average human completion time of \SI{60}{\second}.
We randomize the robot starting position, box placement, objects inside the box, shelf empty spaces, and distractors.

\textit{Activity 5 \textbf{Lay Clothes Out} (Fig.~\ref{fig:pull} Fifth Row)}:
In a bedroom, the robot moves to a wardrobe, opens it (\textbf{ST-1}), picks up a jacket on a hanger (\textbf{ST-2}), lays the jacket on a sofa bed (\textbf{ST-3}), and then returns to close the wardrobe (\textbf{ST-4}).
\bimanual{Bimanual} coordination is the most critical capability for this task.
We collect 98 demonstrations, with an average human completion time of \SI{120}{\second}.
We randomize the robot starting position, clothing placements, and clothing instances.

\subsection{Policy Evaluation Results}
\label{appendix:sec:policy_evaluation_results}
Numerical results from policy evaluation are presented in Tables~\ref{appendix:table:numerical_results_clean_house_after_a_wild_party}, \ref{appendix:table:numerical_results_clean_the_toilet}, \ref{appendix:table:numerical_results_take_trash_outside}, \ref{appendix:table:numerical_results_put_items_onto_shelf}, and \ref{appendix:table:numerical_results_lay_clothes_out}.

\subsection{Simulation Ablation Details}
We design a simulated table-wiping task in OmniGibson~\citep{li2022behaviork} to perform ablation studies. The robot must use whole-body motions to wipe to a target hand position (marked by the yellow hand in Fig.~\ref{fig:sim_ablation}) while maintaining contact with the table surface.
To generate training data, we use cuRobo~\citep{sundaralingam2023curobo0} to produce 100,000 whole-body trajectories, constraining the motion space by locking the mobile base and the first two torso joints.
To isolate the effects of autoregressive whole-body action decoding and multi-modal observation attention, we replace camera input with a goal position, treated as a separate observation modality alongside robot proprioception.

\subsection{User Study Details}
\label{appendix:sec:user_study_details}
\begin{wrapfigure}[12]{r}{0.5\textwidth}
    \vspace{-1.5cm}
    \centering
    \includegraphics[width=0.5\textwidth]{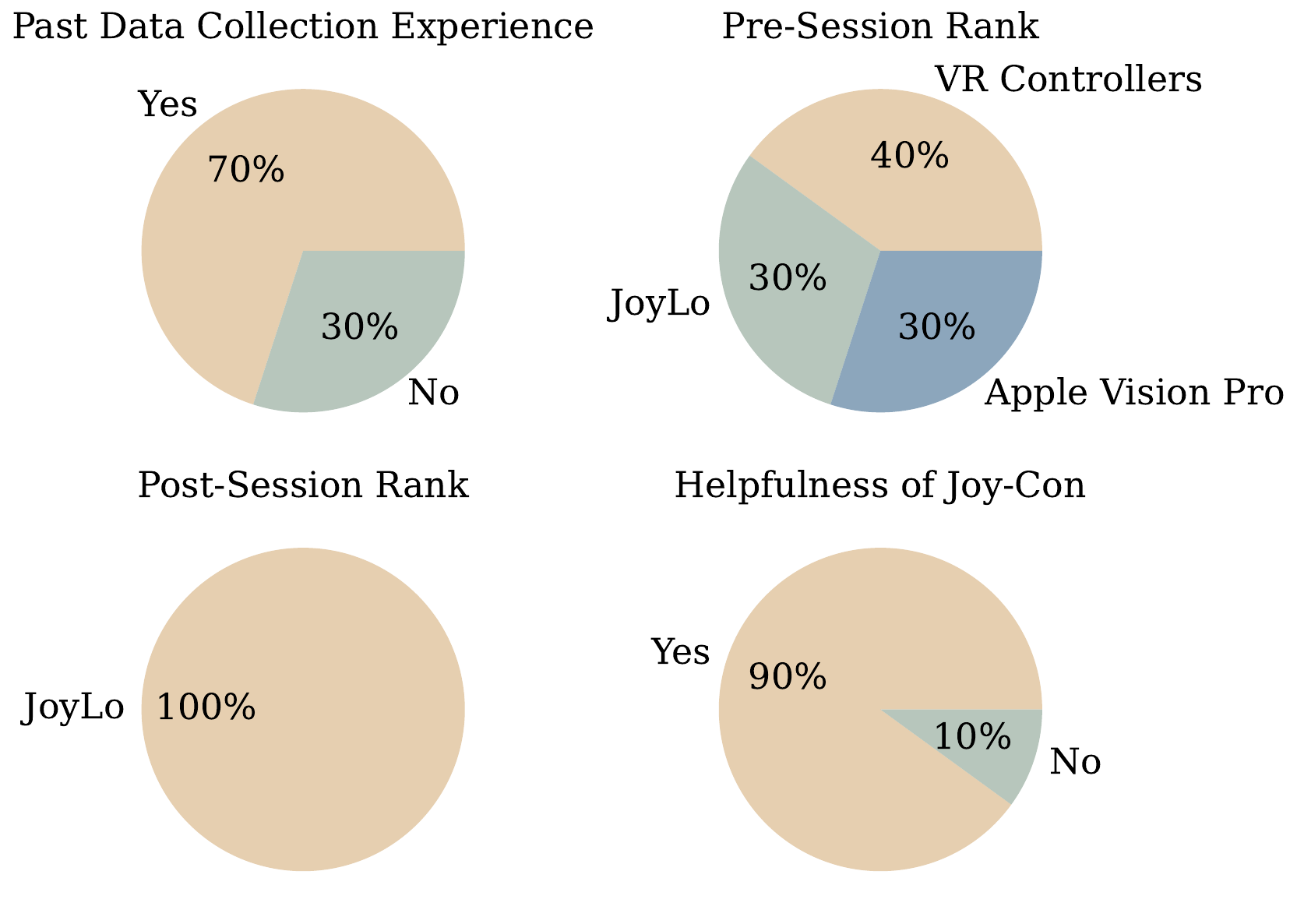}
    \caption{\textbf{Participant demographics and questionnaire results.}}
    \label{appendix:fig:user_study_questionnaires}
\end{wrapfigure}

\begin{figure*}[h]
    \centering
    \includegraphics[width=\textwidth]{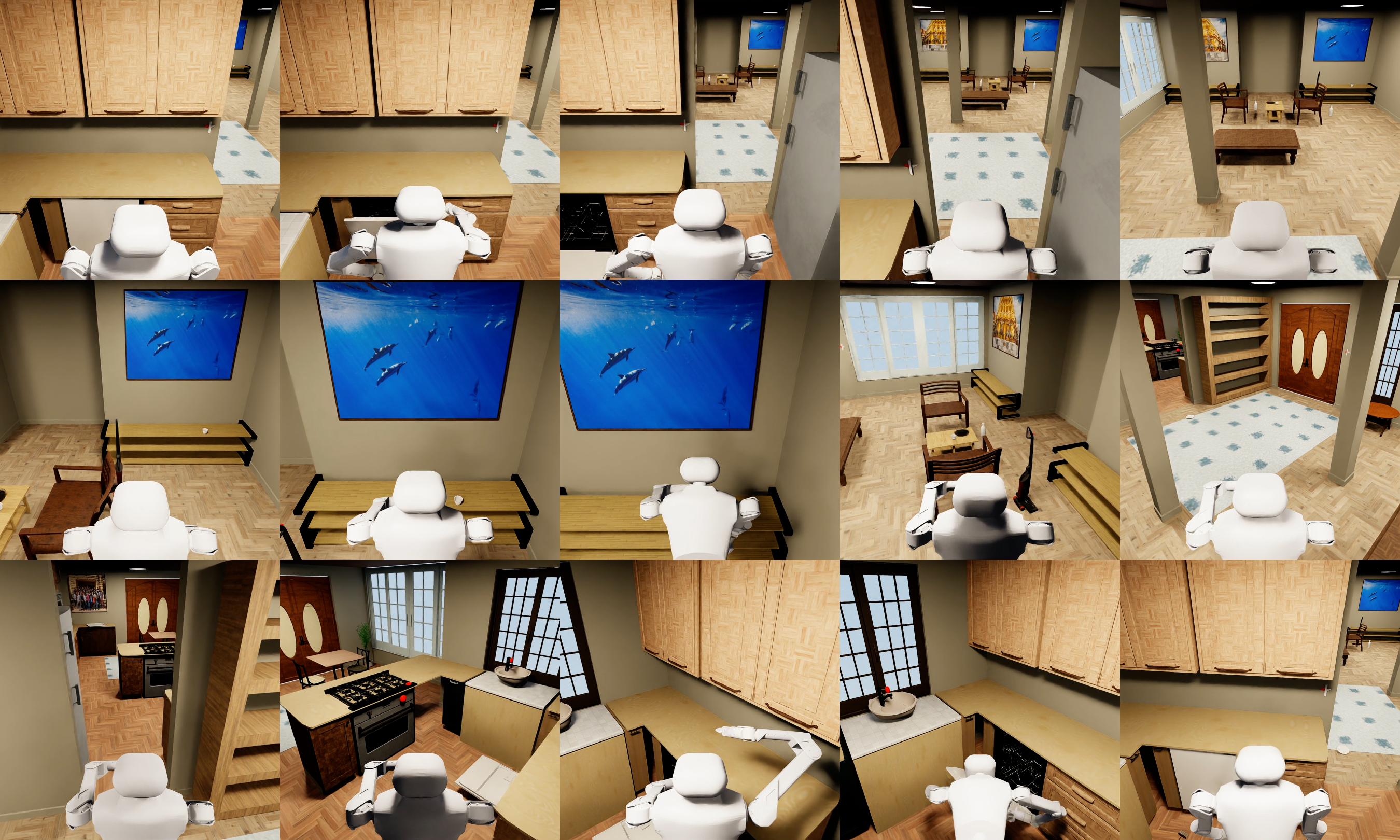}
    \caption{\textbf{Successful task completion by a participant.} The robot navigates to a dishwasher and opens it, moves to a table to collect teacups, returns to the dishwasher, places the teacups inside, and closes it.}
    \label{appendix:fig:user_study_example}
\end{figure*}

As described in Sec.~\ref{sec:exp:user_study}, we conducted a user study with 10 participants to compare \interfacename against two alternative interfaces: VR controllers~\citep{dass2024telemoma} and Apple Vision Pro~\citep{cheng2024opentelevision,park2024avp}.  
The study was conducted in the OmniGibson simulator~\citep{li2022behaviork} on the task ``clean house after a wild party.''  
To provide equal depth perception, participants wore a Meta Quest 3 headset while using both \interfacename and VR controllers.  
To eliminate bias, participants were exposed to the three interfaces in a randomized order. Each participant had a 10-minute practice session for each interface before beginning the formal evaluation.  
A successful task rollout is shown in Fig.~\ref{appendix:fig:user_study_example}.

\begin{wrapfigure}[13]{r}{0.5\textwidth}
    \vspace{-0cm}
    \centering
    \includegraphics[width=0.5\textwidth]{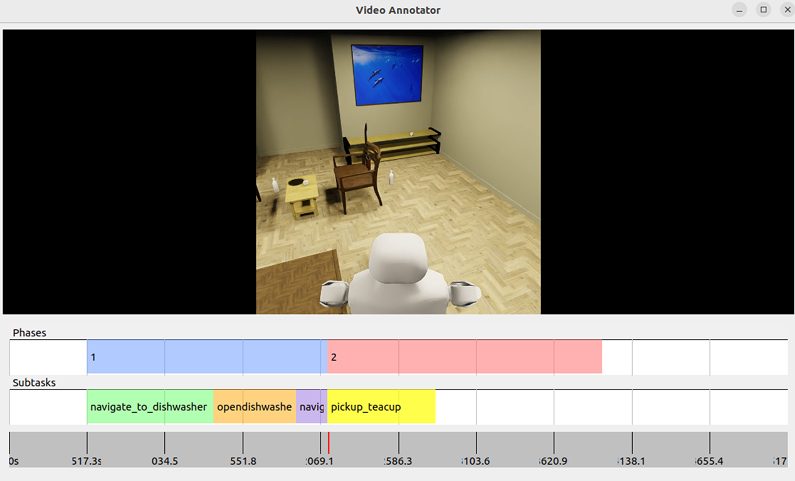}
    \caption{\textbf{GUI for annotating user study rollouts.}}
    \label{appendix:fig:user_study_gui}
\end{wrapfigure}

After the sessions, rollouts were manually segmented, and task and sub-task completions were annotated using a GUI (Fig.~\ref{appendix:fig:user_study_gui}).  
For VR controllers and Apple Vision Pro, which use inverse kinematics (IK) based on end-effector poses, singular configurations were identified when the Jacobian matrix's condition number exceeded a set threshold.  
For \interfacename, which directly controls joints, excessive joint velocities were used as an indicator of singular or near-singular configurations.  
The post-session survey questions sent to participants are listed below:

\begin{enumerate}
\setlength{\itemsep}{0cm}
    \item[\q{1}:]{Do you have prior data collection experience in robot learning? [Yes/No]}
    \item[\q{2}:]{Before the session, which device did you expect to be the most user-friendly? [VR/Apple Vision Pro/JoyLo]}
    \item[\q{3}:]{After the session, which device did you find to be the most user-friendly? [VR/Apple Vision Pro/JoyLo]}
    \item[\q{4}:]{Did physically holding \interfacename arms help with data collection? [Yes/No]}
    \item[\q{5}:]{Did using thumbsticks for torso and mobile base movement improve control? [Yes/No]}
\end{enumerate}

\begin{table}[ht]
\caption{\textbf{Numerical evaluation results for the task ``clean house after a wild party.''} Success rates are shown as percentages. Values in parentheses indicate the number of successful trials out of the total trials.}
\label{appendix:table:numerical_results_clean_house_after_a_wild_party}
\centering
    \resizebox{\linewidth}{!}{
\begin{tabular}{@{}rcccccccc@{}}
\toprule
        & \textbf{ET}                                                     & \textbf{ST-1}                                                             & \textbf{ST-2}                                                            & \textbf{ST-3}                                                            & \textbf{ST-4}                                                            & \textbf{ST-5}                                                            & \textbf{ST-6}                                                            & \textbf{Safety Violations} \\ \midrule
Human Teleop.   & \begin{tabular}[c]{@{}c@{}}68\%\\ (50/73)\end{tabular} & \begin{tabular}[c]{@{}c@{}}100\%\\ (73/73)\end{tabular}          & \begin{tabular}[c]{@{}c@{}}93\%\\ (69/74)\end{tabular}          & \begin{tabular}[c]{@{}c@{}}100\%\\ (69/69)\end{tabular}         & \begin{tabular}[c]{@{}c@{}}89\%\\ (64/72)\end{tabular}          & \begin{tabular}[c]{@{}c@{}}94\%\\ (60/64)\end{tabular}          & \begin{tabular}[c]{@{}c@{}}88\%\\ (53/60)\end{tabular}          & N/A               \\
Ours & \begin{tabular}[c]{@{}c@{}}\bestscore{40\%}\\ \bestscore{(6/15)}\end{tabular}  & \begin{tabular}[c]{@{}c@{}}\bestscore{100\%}\\ \bestscore{(15/15)}\end{tabular} & \begin{tabular}[c]{@{}c@{}}\bestscore{80\%}\\ \bestscore{(12/15)}\end{tabular} & \begin{tabular}[c]{@{}c@{}}\bestscore{80\%}\\ \bestscore{(12/15)}\end{tabular} & \begin{tabular}[c]{@{}c@{}}\bestscore{73\%}\\ \bestscore{(11/15)}\end{tabular} & \begin{tabular}[c]{@{}c@{}}\bestscore{93\%}\\ \bestscore{(14/15)}\end{tabular} & \begin{tabular}[c]{@{}c@{}}\bestscore{93\%}\\ \bestscore{(14/15)}\end{tabular} & \bestscore{0}        \\
DP3~\citep{ze20243d}     & \begin{tabular}[c]{@{}c@{}}0\%\\ (0/15)\end{tabular}   & \begin{tabular}[c]{@{}c@{}}80\%\\ (12/15)\end{tabular}           & \begin{tabular}[c]{@{}c@{}}7\%\\ (1/15)\end{tabular}            & \begin{tabular}[c]{@{}c@{}}27\%\\ (4 / 15)\end{tabular}         & \begin{tabular}[c]{@{}c@{}}7\%\\ (1/15)\end{tabular}            & \begin{tabular}[c]{@{}c@{}}33\%\\ (5/15)\end{tabular}           & \begin{tabular}[c]{@{}c@{}}40\%\\ (6/15)\end{tabular}           & 13                \\
RGB-DP~\citep{chi2023diffusion}  & \begin{tabular}[c]{@{}c@{}}0\%\\ (0/15)\end{tabular}   & \begin{tabular}[c]{@{}c@{}}93\%\\ (14/15)\end{tabular}           & \begin{tabular}[c]{@{}c@{}}0\%\\ (0/15)\end{tabular}            & \begin{tabular}[c]{@{}c@{}}0\%\\ (0/15)\end{tabular}            & \begin{tabular}[c]{@{}c@{}}7\%\\ (1/15)\end{tabular}            & \begin{tabular}[c]{@{}c@{}}7\%\\ (1/15)\end{tabular}            & \begin{tabular}[c]{@{}c@{}}20\%\\ (3/15)\end{tabular}           & 2                 \\
ACT~\citep{zhao2023learning}  & \begin{tabular}[c]{@{}c@{}}0\%\\ (0/15)\end{tabular}   & \begin{tabular}[c]{@{}c@{}}80\%\\ (12/15)\end{tabular}           & \begin{tabular}[c]{@{}c@{}}0\%\\ (0/15)\end{tabular}            & \begin{tabular}[c]{@{}c@{}}0\%\\ (0/15)\end{tabular}            & \begin{tabular}[c]{@{}c@{}}0\%\\ (0/15)\end{tabular}            & \begin{tabular}[c]{@{}c@{}}0\%\\ (0/15)\end{tabular}            & \begin{tabular}[c]{@{}c@{}}0\%\\ (0/15)\end{tabular}           & 2                 \\ \bottomrule
\end{tabular}}
\end{table}

\begin{table}[ht]
\caption{\textbf{Numerical evaluation results for the task ``clean the toilet.''} Success rates are shown as percentages. Values in parentheses indicate the number of successful trials out of the total trials.}
\label{appendix:table:numerical_results_clean_the_toilet}
\centering
    \resizebox{\linewidth}{!}{
\begin{tabular}{@{}rcccccccc@{}}
\toprule
        & \textbf{ET}                                                     & \textbf{ST-1}                                                             & \textbf{ST-2}                                                            & \textbf{ST-3}                                                            & \textbf{ST-4}                                                            & \textbf{ST-5}                                                            & \textbf{ST-6}                                                            & \textbf{Safety Violations} \\ \midrule
Human Teleop.   & \begin{tabular}[c]{@{}c@{}}61\%\\ (100/164)\end{tabular} & \begin{tabular}[c]{@{}c@{}}91\%\\ (150/164)\end{tabular} & \begin{tabular}[c]{@{}c@{}}72\%\\ (106/148)\end{tabular} & \begin{tabular}[c]{@{}c@{}}99\%\\ (104/105)\end{tabular} & \begin{tabular}[c]{@{}c@{}}100\%\\ (103/103)\end{tabular} & \begin{tabular}[c]{@{}c@{}}98\%\\ (102/104)\end{tabular} & \begin{tabular}[c]{@{}c@{}}98\%\\ (100/102)\end{tabular} & N/A               \\
Ours & \begin{tabular}[c]{@{}c@{}}\bestscore{53\%}\\ \bestscore{(8/15)}\end{tabular}    & \begin{tabular}[c]{@{}c@{}}\bestscore{100\%}\\ \bestscore{(15/15)}\end{tabular}  & \begin{tabular}[c]{@{}c@{}}\bestscore{80\%}\\ \bestscore{(12/15)}\end{tabular}   & \begin{tabular}[c]{@{}c@{}}\bestscore{100\%}\\ \bestscore{(15/15)}\end{tabular}  & \begin{tabular}[c]{@{}c@{}}\bestscore{100\%}\\ \bestscore{(15/15)}\end{tabular}   & \begin{tabular}[c]{@{}c@{}}\bestscore{100\%}\\ \bestscore{(15/15)}\end{tabular}  & \begin{tabular}[c]{@{}c@{}}\bestscore{73\%}\\ \bestscore{(11/15)}\end{tabular}   & \bestscore{0}                 \\
DP3~\citep{ze20243d}     & \begin{tabular}[c]{@{}c@{}}0\%\\ (0/15)\end{tabular}     & \begin{tabular}[c]{@{}c@{}}\bestscore{100\%}\\ \bestscore{(15/15)}\end{tabular}  & \begin{tabular}[c]{@{}c@{}}47\%\\ (7/15)\end{tabular}    & \begin{tabular}[c]{@{}c@{}}93\%\\ (14/15)\end{tabular}   & \begin{tabular}[c]{@{}c@{}}0\%\\ (0/15)\end{tabular}      & \begin{tabular}[c]{@{}c@{}}13\%\\ (2/15)\end{tabular}    & \begin{tabular}[c]{@{}c@{}}0\%\\ (0/15)\end{tabular}     & \bestscore{0}                 \\
RGB-DP~\citep{chi2023diffusion}  & \begin{tabular}[c]{@{}c@{}}0\%\\ (0/15)\end{tabular}     & \begin{tabular}[c]{@{}c@{}}93\%\\ (14/15)\end{tabular}   & \begin{tabular}[c]{@{}c@{}}13\%\\ (2/15)\end{tabular}    & \begin{tabular}[c]{@{}c@{}}7\%\\ (1/15)\end{tabular}     & \begin{tabular}[c]{@{}c@{}}7\%\\ (1/15)\end{tabular}      & \begin{tabular}[c]{@{}c@{}}0\%\\ (0/15)\end{tabular}     & \begin{tabular}[c]{@{}c@{}}20\%\\ (3/15)\end{tabular}    & 2                 \\
ACT~\citep{zhao2023learning}  & \begin{tabular}[c]{@{}c@{}}0\%\\ (0/15)\end{tabular}     & \begin{tabular}[c]{@{}c@{}}20\%\\ (3/15)\end{tabular}   & \begin{tabular}[c]{@{}c@{}}0\%\\ (0/15)\end{tabular}    & \begin{tabular}[c]{@{}c@{}}0\%\\ (0/15)\end{tabular}     & \begin{tabular}[c]{@{}c@{}}0\%\\ (0/15)\end{tabular}      & \begin{tabular}[c]{@{}c@{}}0\%\\ (0/15)\end{tabular}     & \begin{tabular}[c]{@{}c@{}}0\%\\ (0/15)\end{tabular}    & \bestscore{0}                  \\ \bottomrule
\end{tabular}}
\end{table}

\begin{table}[ht]
\caption{\textbf{Numerical evaluation results for the task ``take trash outside.''} Success rates are shown as percentages. Values in parentheses indicate the number of successful trials out of the total trials.}
\label{appendix:table:numerical_results_take_trash_outside}
\centering
\begin{tabular}{@{}rcccccc@{}}
\toprule
        & \textbf{ET}                                                     & \textbf{ST-1}                                                             & \textbf{ST-2}                                                            & \textbf{ST-3}                                                            & \textbf{ST-4}                                                            & \textbf{Safety Violations} \\ \midrule
Human Teleop.   & \begin{tabular}[c]{@{}c@{}}76\%\\ (96/127)\end{tabular} & \begin{tabular}[c]{@{}c@{}}91\%\\ (116/128)\end{tabular} & \begin{tabular}[c]{@{}c@{}}100\%\\ (124/124)\end{tabular} & \begin{tabular}[c]{@{}c@{}}85\%\\ (106/125)\end{tabular} & \begin{tabular}[c]{@{}c@{}}100\%\\ (115/115)\end{tabular} & N/A               \\
Ours & \begin{tabular}[c]{@{}c@{}}\bestscore{53\%}\\ \bestscore{(8/15)}\end{tabular}   & \begin{tabular}[c]{@{}c@{}}\bestscore{80\%}\\ \bestscore{(12/15)}\end{tabular}   & \begin{tabular}[c]{@{}c@{}}\bestscore{100\%}\\ \bestscore{(15/15)}\end{tabular}   & \begin{tabular}[c]{@{}c@{}}\bestscore{87\%}\\ \bestscore{(13/15)}\end{tabular}   & \begin{tabular}[c]{@{}c@{}}\bestscore{87\%}\\ \bestscore{(13/15)}\end{tabular}    & \bestscore{1}                 \\
DP3~\citep{ze20243d}     & \begin{tabular}[c]{@{}c@{}}0\%\\ (0/15)\end{tabular}    & \begin{tabular}[c]{@{}c@{}}60\%\\ (9/15)\end{tabular}    & \begin{tabular}[c]{@{}c@{}}53\%\\ (8/15)\end{tabular}     & \begin{tabular}[c]{@{}c@{}}20\%\\ (3/15)\end{tabular}    & \begin{tabular}[c]{@{}c@{}}7\%\\ (1/15)\end{tabular}      & 9                 \\
RGB-DP~\citep{chi2023diffusion}  & \begin{tabular}[c]{@{}c@{}}0\%\\ (0/15)\end{tabular}    & \begin{tabular}[c]{@{}c@{}}20\%\\ (3/15)\end{tabular}    & \begin{tabular}[c]{@{}c@{}}7\%\\ (1/15)\end{tabular}      & \begin{tabular}[c]{@{}c@{}}7\%\\ (1/15)\end{tabular}     & \begin{tabular}[c]{@{}c@{}}7\%\\ (1/15)\end{tabular}      & 3                 \\
ACT~\citep{zhao2023learning}  & \begin{tabular}[c]{@{}c@{}}0\%\\ (0/15)\end{tabular}    & \begin{tabular}[c]{@{}c@{}}0\%\\ (0/15)\end{tabular}    & \begin{tabular}[c]{@{}c@{}}0\%\\ (0/15)\end{tabular}      & \begin{tabular}[c]{@{}c@{}}0\%\\ (0/15)\end{tabular}     & \begin{tabular}[c]{@{}c@{}}0\%\\ (0/15)\end{tabular}      & 5                 \\ \bottomrule
\end{tabular}
\end{table}

\begin{table*}[ht]
\caption{\textbf{Numerical evaluation results for the task ``put items onto shelf.''} Success rates are shown as percentages. Values in parentheses indicate the number of successful trials out of the total trials.}
\label{appendix:table:numerical_results_put_items_onto_shelf}
\centering
\begin{tabular}{@{}rcccc@{}}
\toprule
        & \textbf{ET}                                                     & \textbf{ST-1}                                                             & \textbf{ST-2}                                                           & \textbf{Safety Violations} \\ \midrule
Human Teleop.                                                                      & \begin{tabular}[c]{@{}c@{}}89\%\\ (93/104)\end{tabular} & \begin{tabular}[c]{@{}c@{}}90\%\\ (94/104)\end{tabular} & \begin{tabular}[c]{@{}c@{}}100\%\\ (93/93)\end{tabular} & N/A               \\
Ours                                                                    & \begin{tabular}[c]{@{}c@{}}\bestscore{93\%}\\ \bestscore{(14/15)}\end{tabular}  & \begin{tabular}[c]{@{}c@{}}\bestscore{93\%}\\ \bestscore{(14/15)}\end{tabular}  & \begin{tabular}[c]{@{}c@{}}\bestscore{100\%}\\ \bestscore{(15/15)}\end{tabular} & \bestscore{0}                 \\
DP3~\citep{ze20243d}                                                                        & \begin{tabular}[c]{@{}c@{}}20\%\\ (3/15)\end{tabular}   & \begin{tabular}[c]{@{}c@{}}27\%\\ (4/15)\end{tabular}   & \begin{tabular}[c]{@{}c@{}}47\%\\ (7/15)\end{tabular}   & \bestscore{0}                 \\
RGB-DP~\citep{chi2023diffusion}                                                                     & \begin{tabular}[c]{@{}c@{}}13\%\\ (2/15)\end{tabular}   & \begin{tabular}[c]{@{}c@{}}20\%\\ (3/15)\end{tabular}   & \begin{tabular}[c]{@{}c@{}}40\%\\ (6/15)\end{tabular}   & \bestscore{0}                 \\
ACT~\citep{zhao2023learning}                                                                     & \begin{tabular}[c]{@{}c@{}}0\%\\ (0/15)\end{tabular}   & \begin{tabular}[c]{@{}c@{}}0\%\\ (0/15)\end{tabular}   & \begin{tabular}[c]{@{}c@{}}33\%\\ (5/15)\end{tabular}   & 1                 \\
Ours w/o W.B. Action Denoising   & \begin{tabular}[c]{@{}c@{}}40\%\\ (6/15)\end{tabular}   & \begin{tabular}[c]{@{}c@{}}40\%\\ (6/15)\end{tabular}   & \begin{tabular}[c]{@{}c@{}}60\%\\ (9/15)\end{tabular}   & \bestscore{0}                 \\
Ours w/o Multi-Modal Obs. Attn. & \begin{tabular}[c]{@{}c@{}}13\%\\ (2/15)\end{tabular}   & \begin{tabular}[c]{@{}c@{}}33\%\\ (5/15)\end{tabular}   & \begin{tabular}[c]{@{}c@{}}40\%\\ (6/15)\end{tabular}   & \bestscore{0}                 \\ \bottomrule
\end{tabular}
\end{table*}

\begin{table}[ht]
\caption{\textbf{Numerical evaluation results for the task ``lay clothes out.''} Success rates are shown as percentages. Values in parentheses indicate the number of successful trials out of the total trials.}
\label{appendix:table:numerical_results_lay_clothes_out}
\centering
    \resizebox{\linewidth}{!}{
\begin{tabular}{@{}rcccccc@{}}
\toprule
        & \textbf{ET}                                                     & \textbf{ST-1}                                                             & \textbf{ST-2}                                                            & \textbf{ST-3}                                                            & \textbf{ST-4}                                                            & \textbf{Safety Violations} \\ \midrule
Human Teleop.                                                                      & \begin{tabular}[c]{@{}c@{}}50\%\\ (54/108)\end{tabular} & \begin{tabular}[c]{@{}c@{}}56\%\\ (60/108)\end{tabular} & \begin{tabular}[c]{@{}c@{}}93\%\\ (56/60)\end{tabular} & \begin{tabular}[c]{@{}c@{}}96\%\\ (54/56)\end{tabular} & \begin{tabular}[c]{@{}c@{}}100\%\\ (54/54)\end{tabular} & N/A               \\
Ours                                                                       & \begin{tabular}[c]{@{}c@{}}\bestscore{53\%}\\ \bestscore{(8/15)}\end{tabular}   & \begin{tabular}[c]{@{}c@{}}\bestscore{87\%}\\ \bestscore{(13/15)}\end{tabular}  & \begin{tabular}[c]{@{}c@{}}\bestscore{93\%}\\ \bestscore{(14/15)}\end{tabular} & \begin{tabular}[c]{@{}c@{}}\bestscore{80\%}\\ \bestscore{(12/15)}\end{tabular} & \begin{tabular}[c]{@{}c@{}}60\%\\ (9/15)\end{tabular}   & \bestscore{0}                 \\
DP3~\citep{ze20243d}                                                                        & \begin{tabular}[c]{@{}c@{}}0\%\\ (0/15)\end{tabular}    & \begin{tabular}[c]{@{}c@{}}13\%\\ (2/15)\end{tabular}   & \begin{tabular}[c]{@{}c@{}}13\%\\ (2/15)\end{tabular}  & \begin{tabular}[c]{@{}c@{}}27\%\\ (4/15)\end{tabular}  & \begin{tabular}[c]{@{}c@{}}27\%\\ (4/15)\end{tabular}   & 7                 \\
RGB-DP~\citep{chi2023diffusion}                                                                     & \begin{tabular}[c]{@{}c@{}}0\%\\ (0/8)\end{tabular}     & \begin{tabular}[c]{@{}c@{}}13\%\\ (1/8)\end{tabular}    & \begin{tabular}[c]{@{}c@{}}25\%\\ (2/8)\end{tabular}   & \begin{tabular}[c]{@{}c@{}}13\%\\ (1/8)\end{tabular}   & \begin{tabular}[c]{@{}c@{}}13\%\\ (1/8)\end{tabular}    & 3                 \\
ACT~\citep{zhao2023learning}                                                                     & \begin{tabular}[c]{@{}c@{}}0\%\\ (0/15)\end{tabular}     & \begin{tabular}[c]{@{}c@{}}0\%\\ (0/15)\end{tabular}    & \begin{tabular}[c]{@{}c@{}}0\%\\ (0/15)\end{tabular}   & \begin{tabular}[c]{@{}c@{}}0\%\\ (0/15)\end{tabular}   & \begin{tabular}[c]{@{}c@{}}0\%\\ (0/15)\end{tabular}    & 1                 \\
Ours w/o W.B. Action Denoising   & \begin{tabular}[c]{@{}c@{}}13\%\\ (2/15)\end{tabular}   & \begin{tabular}[c]{@{}c@{}}33\%\\ (5/15)\end{tabular}   & \begin{tabular}[c]{@{}c@{}}73\%\\ (11/15)\end{tabular} & \begin{tabular}[c]{@{}c@{}}73\%\\ (11/15)\end{tabular} & \begin{tabular}[c]{@{}c@{}}\bestscore{67\%}\\ \bestscore{(10/15)}\end{tabular}  & \bestscore{0}                 \\
Ours w/o Multi-Modal Obs. Attn. & \begin{tabular}[c]{@{}c@{}}0\%\\ (0/15)\end{tabular}    & \begin{tabular}[c]{@{}c@{}}33\%\\ (5/15)\end{tabular}   & \begin{tabular}[c]{@{}c@{}}40\%\\ (6/15)\end{tabular}  & \begin{tabular}[c]{@{}c@{}}47\%\\ (7/15)\end{tabular}  & \begin{tabular}[c]{@{}c@{}}13\%\\ (2/15)\end{tabular}   & 4                 \\ \bottomrule
\end{tabular}}
\end{table}

\end{document}